\documentclass{article}

\PassOptionsToPackage{table}{xcolor}
\usepackage[preprint]{neurips_2025}

\usepackage[utf8]{inputenc} 
\usepackage[T1]{fontenc}    
\usepackage{hyperref}       
\usepackage{url}            
\usepackage{booktabs}       
\usepackage{amsfonts}       
\usepackage{nicefrac}       
\usepackage{microtype}      

\usepackage{booktabs}
\usepackage{diagbox}  
\usepackage{amsmath}
\usepackage{amssymb}
\usepackage{enumerate}
\usepackage{amsthm}
\usepackage{amsfonts}
\usepackage{ifthen}
\usepackage{graphicx}
\usepackage{subcaption}
\usepackage{cleveref}
\usepackage[table]{xcolor}
\usepackage{multirow}
\usepackage{algorithm,algpseudocode,amsmath}

\newcommand{\bW}{\mathbf{W}}

\newcommand{\be}{\begin{equation}}        
\newcommand{\ee}{\end{equation}}          
\newcommand{\pdr}[2]{\frac{\partial #1}{\partial #2}}    

\newcommand{\norm}[1]{\left\lVert#1\right\rVert}

\newcommand{\floor}[1]{\lfloor #1 \rfloor}

\newcommand{\LN}{\textrm{LN}}
\newcommand{\veps}{\varepsilon}
\newcommand{\bzeta}{\boldsymbol{\zeta}}

\newcommand{\lambdamax}{\lambda_{\textrm{max}}}
\newcommand{\opnorm}[1]{\norm{#1}_{\textrm{op}}}

\newtheorem  {lemma*} {Lemma}

\title{Characterization and Mitigation of Training Instabilities in Microscaling Formats}

\author{%
  Chloe Huangyuan Su\thanks{Equal contribution; correspondence: \href{nikhil\_anand@harvard.edu}{nikhil\_anand@harvard.edu}.} \ \textsuperscript{1,2} \\
  \And
  Mujin Kwun\textsuperscript{1} \\
  \And
  Stephanie Gil\textsuperscript{2} \\
  \And
  Sham Kakade\textsuperscript{1,2} \\
  \And
  Nikhil Anand\footnotemark[1] \ \textsuperscript{1} \\
  \\
  \textsuperscript{1}Kempner Institute for the Study of Natural and Artificial Intelligence, Harvard University \\
  \textsuperscript{2}Department of Computer Science, Harvard University \\
}

\begin{document}

\maketitle

\begin{abstract}
    Training large language models is an expensive, compute-bound process that must be repeated as models scale, algorithms improve, and new data is collected. To address this, next-generation hardware accelerators increasingly support lower-precision arithmetic formats, such as the Microscaling (MX) formats introduced in NVIDIA’s Blackwell architecture. These formats use a shared scale within blocks of parameters to extend representable range and perform forward/backward GEMM operations in reduced precision for efficiency gains. In this work, we investigate the challenges and viability of block-scaled precision formats during model training. Across nearly one thousand language models trained from scratch -- spanning compute budgets from \( 2 \times 10^{17} \) to \( 4.8 \times 10^{19} \) FLOPs and sweeping over a broad range of weight–activation precision combinations -- we consistently observe that training in MX formats exhibits sharp, stochastic instabilities in the loss, particularly at larger compute scales. To explain this phenomenon, we conduct controlled experiments and ablations on a smaller proxy model that exhibits similar behavior as the language model, sweeping across architectural settings, hyperparameters, and precision formats. These experiments motivate a simple model in which multiplicative gradient bias introduced by the quantization of layer-norm affine parameters and a small fraction of activations can trigger runaway divergence. Through \textit{in situ} intervention experiments on our proxy model, we demonstrate that instabilities can be averted or delayed by modifying precision schemes mid-training.  Guided by these findings, we evaluate stabilization strategies in the LLM setting and show that certain hybrid configurations recover performance competitive with full-precision training. We release our code at \href{https://github.com/Hither1/systems-scaling}{https://github.com/Hither1/systems-scaling}.
    
\end{abstract}

\section{Introduction\label{sec:intro}}

Large language models (LLMs) have dramatically improved in capabilities in recent years, largely driven by scaling their capacity and the quantity of training data~\citep{kaplan2020scaling, openai2025gpt45, deepmind2025gemini25, anthropic2025claude4, grattafiori2024llama}. For instance, training the Llama 3.1 405B model required more than $10^{25}$ FLOPs and utilized up to 16,000 H100 GPUs~\citep{grattafiori2024llama}. Scaling these models involves not only the initial, compute-intensive pretraining phase but also frequent retraining as new data becomes available, algorithms evolve, or architectural modifications are introduced, as well as post-training protocols that prepare the model for inference/deployment. Pretraining itself also adds the additional challenge of carefully tuned hyperparameter settings, such as learning rate schedules that decay toward zero, making it difficult to resume a completed run without resetting the optimization state. As a result, even small updates -- such as incorporating new data or minor architectural changes -- often require restarting training from early checkpoints, further compounding compute demands~\citep{ibrahim2024simple}. 

To reduce these computational burdens, recent hardware advancements have introduced native support for lower-precision computations, such as FP8 training in NVIDIA H100 GPUs \citep{micikevicius2022fp8formatsdeeplearning}. Upcoming hardware accelerators powered by NVIDIA's Blackwell architecture will further extend these capabilities with standardized, shared-scale Microscaling (MX) formats like MXFP8 and MXFP6~\citep{website3}. These formats store a per-block shared scale, which expands the effective dynamic range with minimal memory overhead, while simultaneously enabling GEMMs at lower precision~\citep{rouhani2023microscaling, darvish2023shared}. While model pretraining is typically done in 16 or 32-bit precision, some quantization schemes are already seeing industry adoption; for example, DeepSeek-V3 employs tile-wise FP8 quantization within large tensors~\citep{liu2024deepseek}, while Cohere’s Command A model was trained in FP8 while reserving higher-precision operations for, e.g., activation functions and attention mechanisms~\citep{cohere2025commandaenterprisereadylarge}.  At an even larger scale, the Llama-4 series of models is reported to have been pretrained in FP8 precision across nearly 32,000 GPUs~\citep{llama4}. On the deployment side, workflows like quantization-aware training (QAT) and mixed-precision fine-tuning further underscore that understanding low-precision training dynamics is essential throughout a model’s lifecycle~\citep{jacob2017quantizationtrainingneuralnetworks, Abdolrashidi_2021, shao2024omniquantomnidirectionallycalibratedquantization}.

There are two primary challenges that accompany the adoption of low-precision formats for training. First, there is a potential performance tradeoff, where reducing precision may result in degradation or "plateauing" of training loss and downstream accuracy, which can be characterized through scaling laws that account for both compute and precision~\citep{kumar2024scaling}. Second, instabilities may arise in training, often manifesting as abrupt spikes in the loss curve that disrupt convergence~\citep{fishman2024scaling, lee2025fp8againquantifyingreduced}.  When these instabilities push optimization into regions from which recovery is impossible, they obstruct our ability to extract valid scaling laws, making it impossible to even assess the tradeoffs introduced by low-precision training.

In this work, we set out to understand the training dynamics of low-precision MX precision formats and extract valid scaling laws to identify format prescriptions for language model training on next-generation hardware. However, consistent with prior observations by \citet{fishman2024scaling, lee2025fp8againquantifyingreduced}, we found that training frequently became unstable -- particularly for larger, compute-intensive models. In contrast to earlier work, these instabilities appeared more pervasive, emerging across a broad range of activation functions, model scales, MX formats, and hyperparameter settings.

Because large-scale language model sweeps are computationally intensive and involve many entangled components, we turn to a controlled synthetic setting to better understand the origin of these failures. Specifically, we employ a residual multi-layer perceptron (MLP) student-teacher model trained on random (Gaussian) inputs, which enables fine-grained ablations over architecture and optimization. This proxy model allows us to isolate the effects of low-precision arithmetic and identify conditions under which training becomes unstable.

In this setting, we observe two qualitatively distinct modes of instability. The first is ``expected" and stems from stochastic optimization dynamics: for instance, aggressive learning rates can amplify small incorrect gradient updates, resulting in loss spikes and divergence. The second, perhaps more interesting, mode is directly induced by low-precision arithmetic: even under stable hyperparameters, quantization introduces systematic gradient bias that can accumulate and destabilize training.  

To better understand the second failure mode, we perform ablations across format configurations, quantization schemes (e.g., forward-only vs. full quantization), and activation functions, and analyze their effects on stability. Our results support a phenomenological explanation: low-precision instabilities primarily arise from systematic bias in gradient estimates introduced by quantization. We find that a key driver of this bias is the quantization of the layer normalization affine parameters, whose values often become tightly clustered over the course of training. When the values within a block converge too closely, division by the  shared block scale can clamp all values in that block to the largest representable number, destabilizing training. We verify that this mechanism is not limited to synthetic settings but also emerges in  language model setting.

With these insights, we propose two fixes to mitigate the instabilities in the language model setting, which involve keeping the MX activations in higher precision or only including quantization in the forward pass.  Applying these fixes, we then train another set of language models in both MXFP8 precision schemes (\texttt{E4M3}, \texttt{E5M2}) and fit valid empirical scaling laws in both cases.

Main contributions:

\begin{itemize}
    \item \textbf{MX sweeps:} We perform extensive language model pretraining sweeps across weight and activation MX precision formats, and consistently observe training instabilities, particularly in larger models that are trained for longer. While prior work has reported instabilities in low-precision training~\citep{fishman2024scaling, lee2025fp8againquantifyingreduced}, we find them in shared-scale (MX) formats and occurring across a wide range of quantization configurations and activation functions.
    \item \textbf{Mechanistic analysis:} Through a suite of ablations in a simplified student–teacher MLP model, we isolate two modes of instability: one arising from stochastic optimization dynamics, and another induced  by quantization noise. For the second case we show that the gradient becomes systematically biased, derive a norm-based condition that predicts when this bias will dominate, and trace its origin to the data being quantized: most of the layer-norm affine weights (and, to a lesser extent, roughly $\sim 1$\% of activations) are clustered such that MX block-scaling forces them into the same quantization bin.
    \item \textbf{Effective mitigations:} We evaluate two stabilization strategies in full-scale language model training: (1) disabling backward-pass quantization and (2) retaining higher-precision activations. Both approaches enable the recovery of valid empirical scaling laws. Notably, we find that MXFP8 weights (\texttt{E4M3}) paired with BF16 activations can match the performance of full bfloat16 baselines, at least up to the model scales we explored. Our work provides both a diagnostic methodology and empirical reference points for evaluating these nascent low-precision training formats.
    \item \textbf{Broader relevance:} While our experiments focus on models trained from scratch, the gradient-bias phenomena we study are intrinsic to block-scaled quantization and may also inform post-training workflows (e.g.\ QAT) or other low-precision scenarios.
\end{itemize}

\section{Related Work \label{sec:related work}} 

\paragraph{Low-Precision Instabilities} Training large Transformer models at scale can reveal instabilities that can disrupt or even halt learning~\citep{liu2024deepseek, chowdhery2022palmscalinglanguagemodeling, dehghani2023scalingvisiontransformers22, zhang2022optopenpretrainedtransformer, molybog2023theoryadaminstabilitylargescale, fishman2024scaling, zoph2022stmoedesigningstabletransferable, ma2025understandingsilentdatacorruption}.  In many cases, these issues are exacerbated -- or directly triggered -- by low-precision quantization.  For example, \citet{fishman2024scaling} demonstrate that FP8 pretraining becomes unstable when combined with the SwiGLU activation function, attributing the issue to an outlier amplification effect that worsens due to progressive weight alignment over the course of training. Similarly, \citet{lee2025fp8againquantifyingreduced} report that approximately 10\% of FP8 runs using the NanoGPT codebase fail to converge, whereas full-precision (BF16) training exhibits no such failures. Other works \citep{sun2024massive, bondarenko2023quantizabletransformersremovingoutliers,xuimproved}, point to activation outliers and gradient norm growth as contributors to these failures while ~\citet{tseng2025trainingllmsmxfp4} proposes a stochastic rounding based algorithm to stabilize training in MXFP4 formats.  Meanwhile, DeepSeek-V3 also attributes certain training failures due to blockwise quantization of activation gradients~\citep{liu2024deepseek}, underscoring the breadth of challenges introduced by quantization schemes.

\citet{DBLP:conf/iclr/WortsmanLXEAACG24} use small-scale proxy models to study training instabilities in the context of growth of output and layer logits. We adopt a similar approach, and use a simplified student-teacher proxy model that replicates low-precision instabilities in LLMs and introduce a phenomenological framework to distinguish between stochastic and quantization-induced failure modes.

\paragraph{Precision Scaling Laws}
In parallel, several works have examined how model performance scales under precision constraints. This includes both quantization-aware training (QAT), which injects quantization noise during the forward pass to prepare models for low-bit inference, and full low-precision training, which applies reduced-precision arithmetic in both passes to accelerate training~\citep{jacob2017quantizationtrainingneuralnetworks, Abdolrashidi_2021, shao2024omniquantomnidirectionallycalibratedquantization, chen2025scalinglawquantizationawaretraining}.

\citet{kumar2024scaling} compare full-precision, QAT, and low-precision training, finding that overtrained models are especially sensitive to quantization and that even 16-bit training may not be optimal. However, their setup restricts quantization to the forward pass to allow for a fair comparison to QAT methods and does not analyze instability dynamics. We evaluate full MX quantization -- including both forward and backward passes-- and propose stabilization techniques that recover valid scaling laws in those lower-precision MX formats.

\citet{liu2025paretoq} explore QAT scaling laws in  low-bit regimes, finding that fine-tuning outperforms both post-training quantization and QAT from scratch, even down to binary and ternary formats. \citet{ouyang2024low} examine how quantization-induced degradation (QiD) varies across training scales, while \citet{dettmers2023case} study inference-time scaling laws, concluding that 4- and 6-bit models often lie on the Pareto frontier of accuracy and efficiency. 

\subsection{Review of MX Formats and Experimental Approach\label{sec:MX_review}}

Microscaling (MX) formats are a class of low-precision numerical representations designed to enhance the efficiency of deep learning models~\citep{ocp_mx, rouhani2023microscaling}. The idea is conceptually simple as described in~\Cref{algo:convert_to_mx}: we represent a block of $k$ values, $\{V_i\}_{i=1}^k$, using a single shared scale factor $X$ and $k$ corresponding low-precision elements $\{P_i\}$ where the $P_i$ are obtained by casting $V_i / X$ to the specified low-precision format\footnote{We present results for a block size $k=32$ to match what will be hardware supported, but we also experimented with different choies of $k$ and did not observe qualitatively different results than those presented.}. The scale $X$ can be calculated using $X = 2^{\floor{\log_2(\max_i(|V_i|))} - e_{\text{max elem}}}$ with $e_{\text{max elem}}$ being the exponent of the largest normal number representable in the chosen element data format. Common element types include 8-bit floating point (FP8) such as \texttt{E4M3} and \texttt{E5M2}, and 6-bit (FP6) formats like \texttt{E2M3} and \texttt{E3M2}, typically utilizing an 8-bit exponent (\texttt{E8M0}) for the shared scale.

In our experiments, we quantize both weights and activations using these MX formats using the MX Pytorch Emulation Library~\citep{mx_library}. This quantization is applied dynamically to the inputs of matrix multiplication operations (e.g., within Linear, MatMul, BMM layers) across both the forward and backward passes, with results dequantized to a higher precision format (e.g., bfloat16) after the operation. We mainly explore various MX configurations, including the aforementioned 6-bit and 8-bit element formats, and scenarios where MX-formatted tensors are mixed with bfloat16 tensors. We defer a more detailed review of the MX scheme to~\Cref{app:review_mx}.

\section{LLM Experiments \label{sec:llm_exp}}  

\subsection{Setup}For our language model experiments, we use OLMo~\citep{groeneveld2024olmo} in combination with the MX PyTorch Emulation Library~\citep{mx_library} to enable training under various low-precision configurations. All language models use the GeLU activation function; full hyperparameter details are provided in \Cref{tab:model-params}.  
We sweep over a wide range of MX precision formats for both weights and activations, including two FP6 variants (\texttt{E3M2}, \texttt{E2M3}), two FP8 variants (\texttt{E4M3}, \texttt{E5M2}), and a bfloat16 baseline. Each configuration applies full quantization to both forward and backward passes to both weights and activations, as implemented in the Microscaling library~\citep{mx_library}. For each format, we train approximately 70 models\footnote{Some runs crashed and could not always be resumed, leading to small differences in number of models trained for each format.} spanning compute budgets from $2 \times 10^{17}$ to $4 \times 10^{19}$ FLOPs. Model sizes range from $\sim$20M to $\sim$1.7B parameters. Token counts are determined using an adapted version of the FLOP accounting code from~\citet{brandfonbrener2024loss}, originally developed for OLMo scaling law experiments. Token-to-parameter ratios in our sweep range from approximately 2 to 156. All models are trained on the Fineweb-Edu dataset~\cite{penedo2024the}, with the longest runs trained on 35B tokens and the shortest runs corresponding to models trained on 301M tokens.

\begin{figure}[ht]
  \centering
  \begin{subfigure}[t]{\textwidth}
    \centering
    \includegraphics[width=0.8\linewidth]{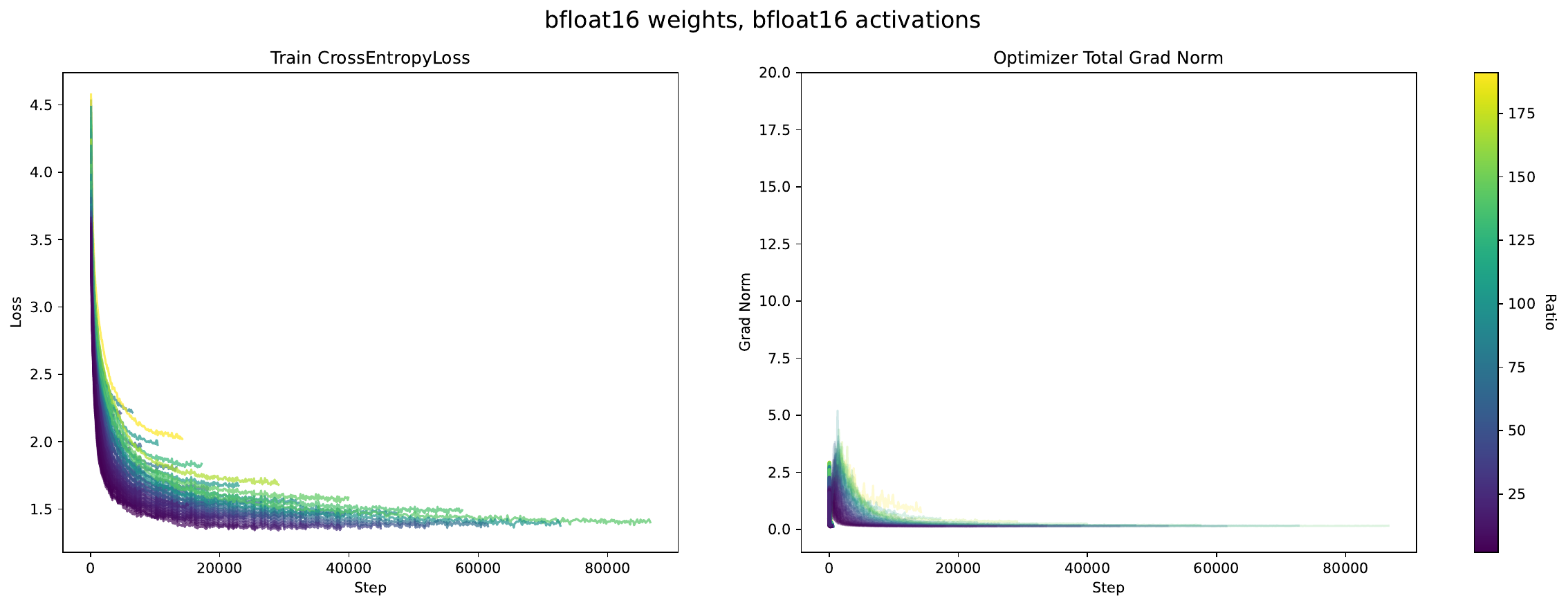} 
    \caption{Train loss and grad norm for weights and activations in bfloat16-bfloat16 format. All runs converge.}
    \label{fig:train_loss_bf16_bf16}
  \end{subfigure}
  \begin{subfigure}[t]{\textwidth}
    \centering
    \includegraphics[width=0.8\linewidth]{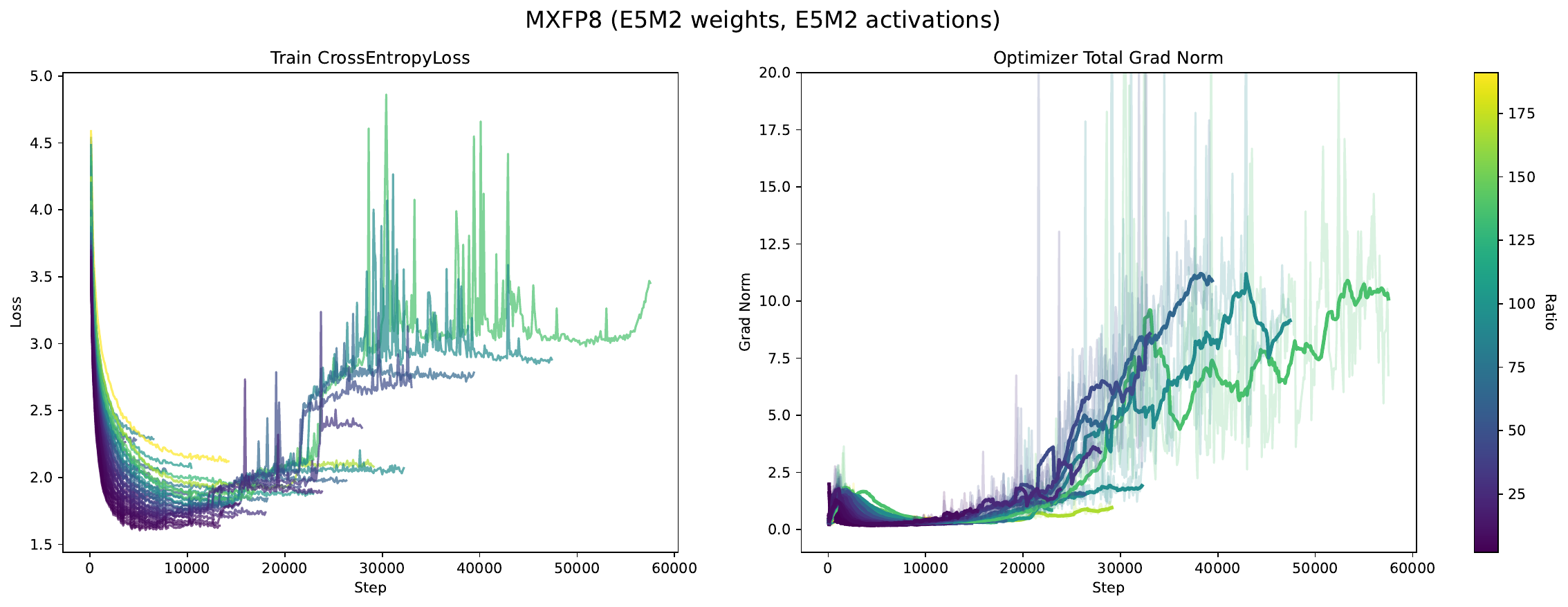} 
    \caption{Train loss and grad norm for weights and activations in MXFP8 \texttt{E5M2}-\texttt{E5M2}. Some runs -- particularly larger models that are trained for longer -- become unstable and never recover.}
    \label{fig:train_loss_fp8_e5m2_fp8_e5m2}
  \end{subfigure}
  \caption{Shows stable (bfloat16) OLMo training runs (top) compared to lower precision (MXFP8 \texttt{E5M2}, bottom). The low precision computations are done in both forward and backward steps, on both weights and activations. Color bar on the right shows the token-to-parameter ratio.}
  \label{fig:scaling-laws-unstable}
\end{figure}

\subsection{Instabilities in Low Precision}
\Cref{fig:train_loss_bf16_bf16} shows the training loss and gradient norm trajectories for bfloat16 models. Training remains stable, with smooth convergence and bounded gradients.  By contrast, \Cref{fig:train_loss_fp8_e5m2_fp8_e5m2} illustrates example instabilities in the MXFP8 \texttt{E5M2}-\texttt{E5M2} weights-activations configuration, where some training runs exhibit sharp upward spikes in loss and large increases in gradient norm magnitude.  We find these instabilities to be universal across other low-precision MX configurations and hyperparameter settings, as documented in Appendix~\ref{app:additional_lang_sweeps}.  We observe the instabilities mainly occur in larger, longer-trained models and that importantly, when training is destabilized, training does not recover, and the loss continues to diverge.  While the loss spikes appear abruptly, the gradient norm typically grows more gradually (bottom right of~\Cref{fig:train_loss_fp8_e5m2_fp8_e5m2}) and fails to decrease over time as seen in stable bfloat16 training. This behavior strongly suggests systematic errors in gradient computation, a point that we will investigate further in subsequent sections.

\section{Synthetic Experiments \label{sec:synthetic}}  

\subsection{Setup} Our language model experiments with OLMo involve many potentially interacting components, and it is challenging to determine exactly where the low-precision failure mode occurs.  To better understand the origins of the instabilities, following the logic of~\citet{DBLP:conf/iclr/WortsmanLXEAACG24}, we develop a small-scale proxy model. Despite its simplicity, this model exhibits many of the same qualitative behaviors seen in large-scale training.

Given an input $x \in \mathbb{R}^{d_{\textrm{model}}}$, we consider a student network composed of $L$ residual layers indexed by $k = 0, \dots, L-1$. The hidden state at each layer is computed as: \be \begin{aligned} A_0 &= x \quad\quad\quad h_k = \bW_k^{(1)} \LN(A_{k-1}) \\
A_{k > 0} &= A_{k-1} + \bW_k^{(2)} \phi(h_k)  \end{aligned}, \label{eq:student_teacher} \ee where $\LN$ denotes layer normalization and $\phi$ is the activation function (e.g., ReLU, GeLU, SwiGLU). Each residual block contains two weight matrices: $\bW_k^{(1)}$ projects to the hidden dimension, and $\bW_k^{(2)}$ projects back to $d_{\textrm{model}}$. By default, the hidden size is set to $4d_{\textrm{model}}$\footnote{In the case of SwiGLU, following~\cite{shazeer2020gluvariantsimprovetransformer} we reduce the hidden dimension from $4d_{\textrm{model}}$ to $\frac{8}{3}d_{\textrm{model}}$ to maintain parity in parameter count.})

The targets are generated by a fixed teacher model whose architecture can be taken to be the same as the student's without the layer normalization (for sweeps where we change the depth and width of the student, we similarly scale the teacher model). A small Gaussian label noise ($\sigma = 10^{-3}$) is added to the outputs. The inputs $x$ are drawn i.i.d. from a standard Gaussian, without cycling, using a fixed seed to ensure consistent batch order.

To isolate the effect of precision, we train two copies of the student model from the same initialization. The first is trained in full precision (FP32). After training, the weights are reset to their initial state and retrained using a low-precision MX format, with quantization applied to both forward and backward passes as described in \Cref{sec:MX_review}. Because the initialization, data, and batch order are identical, any behavioral difference is attributable primarily to the change in numerical precision\footnote{To qualify this statement, we observe that some minor sources non-determinism remain even after controlling for random seed, batch order, model initialization, using deterministic Pytorch kernels, etc. but these effects are small compared to the typical precision scales we work with.}. All models are trained  with an MSE loss.

\paragraph{Hyperparameter choices} A key point is that there are regions in hyperparameter space for which the model in \Cref{eq:student_teacher} will give rise to train instabilities (even in FP32 precision).  This is not necessarily due to the precision scheme chosen, but rather due to the fact that in any SGD method there exists some small probability of taking wrong gradient step(s).  If the size of the steps are large due to, e.g., a high learning rate, this will be visible as a sudden spike in the loss. 

Our goal is to choose hyperparameters and make architectural choices (such as depth and width) in order to move away from these ``expected" instabilities and to ones that could be caused by low precision. For the same reason, we fix a moderately large batch size (2048) throughout to reduce variance in gradient estimates.

\subsection{Sweeping over learning rates and architectures} \paragraph{Learning rates} We begin by sweeping over learning rates $\eta \in (1\times 10^{-5}, 5\times 10^{-5}, 1\times 10^{-4}, 5\times 10^{-4}, 1\times 10^{-3})$ across a range of model depths and widths, in two low precision formats: (1) MXFP8 \texttt{E4M3} in the forward pass and MXFP8 \texttt{E5M2} in the backward pass\footnote{We use this asymmetric format to allow greater dynamic range in the backward pass, following~\citet{micikevicius2022fp8}, and because it exhibited marginally greater stability than using \texttt{E4M3} for both passes.  Our results are not sensitive to this particular choice of low-precision formats.}, and (2) MXFP6 \texttt{E4M3} in both forward and backward passes.  

\begin{figure}[ht]
  \centering
    \includegraphics[width=\linewidth]{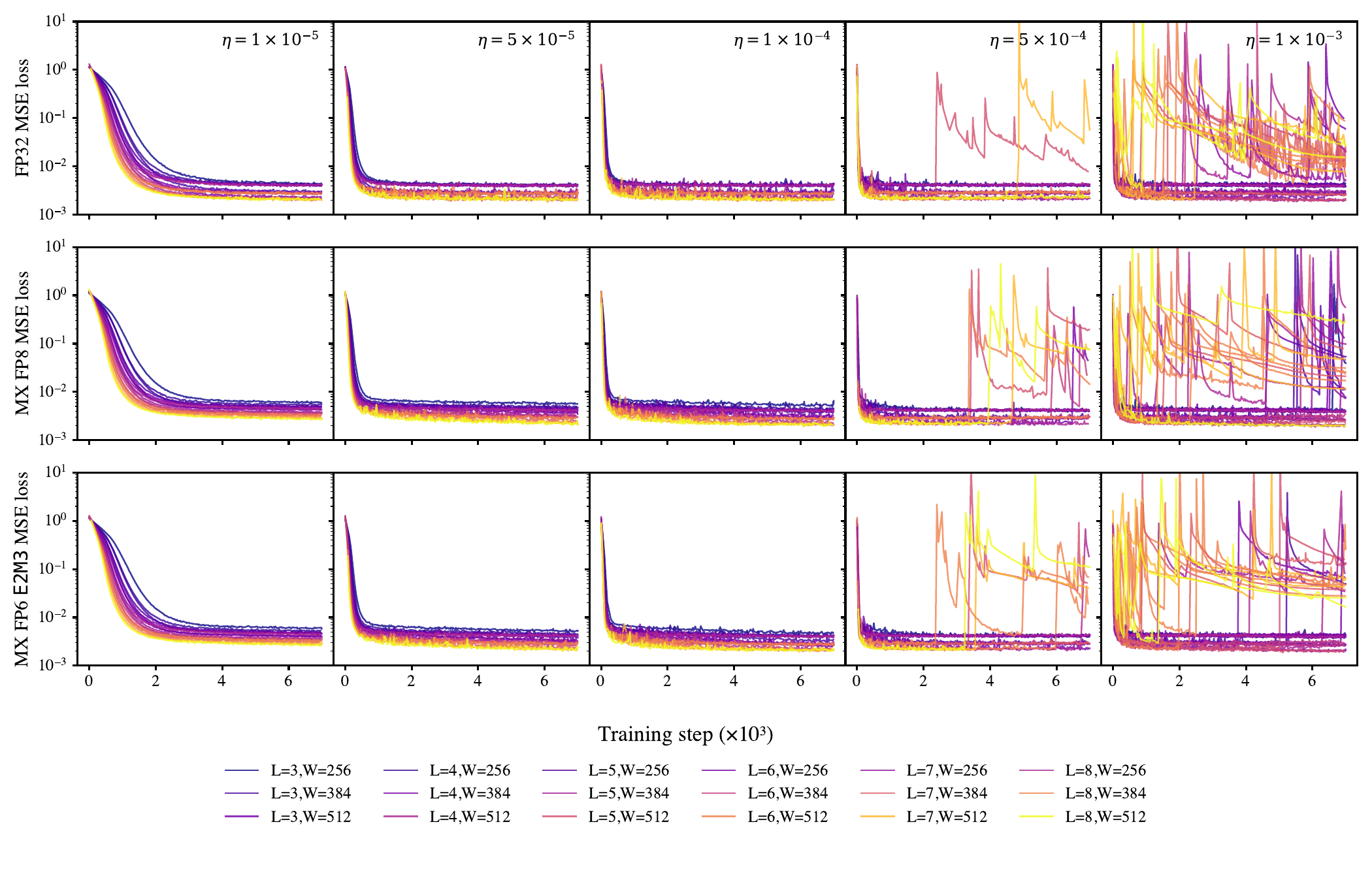} 
  \caption{Comparing FP32 with MXFP6 and MXFP8 formats across different choices for the learning rate.  Color corresponds to model size, determined by the depth $L$ and $d_{\textrm{model}} = D$ on the legend.}
  \label{fig:synthetic_LR_sweep}
\end{figure}

Results from this sweep are shown in \Cref{fig:synthetic_LR_sweep}. We observe the following patterns: for low learning rates $\eta \lesssim 1 \times 10^{-4}$, all precision formats remain stable. At $\eta = 5 \times 10^{-4}$, differences between FP32 and lower-precision formats begin to emerge: FP32 exhibits two unstable runs, while FP8 shows six. At the highest learning rate ($\eta = 1\times 10^{-3}$), instabilities are observed across all formats, with larger models failing earlier in training. Interestingly, we find that recovery from an instability is more rapid in FP32, whereas instability in lower-precision formats--particularly FP6--is often more persistent.

We also experimented with a cosine learning rate schedule that starts at $1\times 10^{-3}$ and decays to $1\times 10^{-5}$ and found that the effect of the schedule was mainly to suppress instabilities at later training times, though we still observe the same differences between high and low precision if the instability does not happen late in training.

\paragraph{Effect of Depth and Width} In Appendix~\ref{app:additional_syn_sweeps}, we fix the learning rate to $\eta = 5 \times 10^{-4}$ and conduct a broader sweep over network depth and width across three MX precision formats. While the number of instability spikes in any individual run is an $O(1)$ integer, aggregating across all runs we observe a general trend: for this fixed learning rate, high-precision training remains stable at larger model sizes than low-precision training.

We find that instability differences between high and low precision seem to occur more frequently in networks of intermediate size, for model dimensions in the range $384 \lesssim d_{\textrm{model}} \lesssim 768$ and depths $3 \lesssim L \lesssim 6$. Intuitively, this makes sense since these models appear to be large enough to exhibit sensitivity to low-precision effects, yet not large enough where overall stochasticity causes generally unstable training at this learning rate. Based on this observation, we fix the learning rate to $\eta = 5 \times 10^{-4}$ in the remainder of our experiments and zoom in on this intermediate model regime.

\subsection{The Effect of Activation Functions and Layernorms} Having identified a hyperparameter regime in which instabilities are more prevalent in low precision than high precision, we next ablate the choice of activation function and the inclusion of layer normalization. In~\Cref{eq:student_teacher}, this corresponds to varying $\phi(\cdot)$ and toggling the presence of $\LN(\cdot)$.

\begin{figure}[ht]
  \centering
  \begin{subfigure}[t]{\textwidth}
    \centering
    \includegraphics[width=\linewidth]{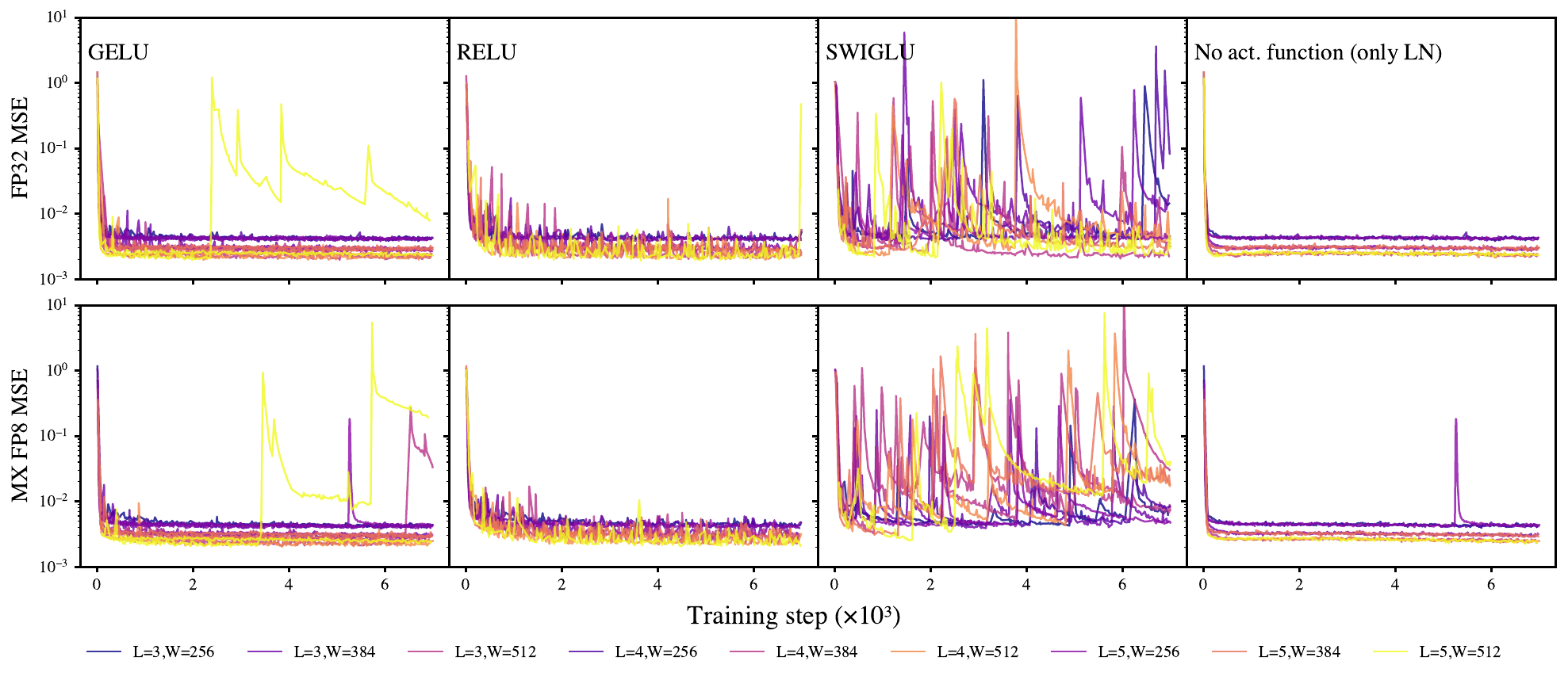} 
    \caption{Loss curves of different activation functions with the inclusion of layernorm.}
    \label{fig:act_sweep_with_ln}
  \end{subfigure}

  \begin{subfigure}[t]{\textwidth}
    \centering
    \includegraphics[width=\linewidth]{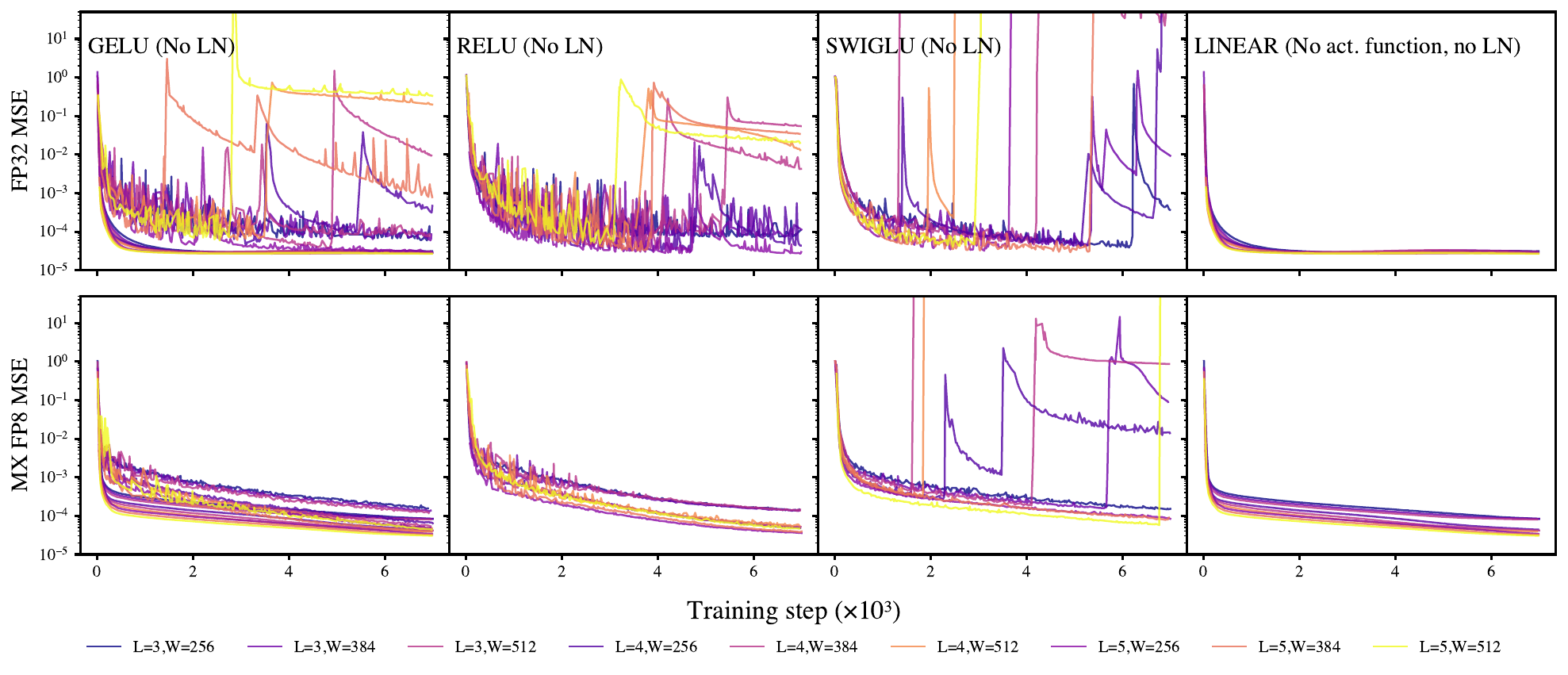} 
    \caption{Loss curves of different activation functions without layernorm.}
    \label{fig:act_sweep_without_ln}
  \end{subfigure}

  \hfill
  \caption{Shows the comparison between full and low precision training across different activation functions, with and without layernorm.}

\end{figure}

In~\Cref{fig:act_sweep_with_ln}, we observe that with layer normalization enabled, both GeLU and SwiGLU activations exhibit instability in low precision, with SwiGLU being significantly more prone to divergence. This is consistent with the findings of~\citet{fishman2024scaling}, though our results show that SwiGLU also destabilizes training in high precision, suggesting that it generally increases training stochasticity at least for this architecture. We observe two instabilities in GeLU under low precision that are absent in high precision, one of which recovers quickly and the other persists.

When layer normalization is removed (\Cref{fig:act_sweep_without_ln}), SwiGLU becomes more stable, while GeLU shows increased variance in high-precision training but not in low precision. For ReLU, the removal of layer norm leads to noisier and sometimes unstable dynamics in high precision but not low precision.  We note that the loss improves with the removal of layernorm; this is expected as the teacher network does not contain a layernorm so that student model is able to more accurately represent its outputs.  At first glance, these results are perplexing -- they suggest that layernorm destabilizes low-precison training while simultaneously stabilizing high-precision training.  We will return to this point in~\Cref{sec:noise_cause} when we explicate the subtleties of layernorms in block scaling formats.

\section{Multiplicative Noise \label{sec:theory}} Our synthetic experiments reveal that training instabilities in low-precision settings can arise from both stochastic optimization effects and quantization-induced bias. These failures appear to result from a complex interplay between architectural choices, activation functions, layer normalization, and learning rate. One hypothesis, motivated by the growth of the gradient norm in~\Cref{fig:scaling-laws-unstable}, is that lower precision is systematically biasing the gradient.  In this section, we examine this hypothesis through a multiplicative noise model and show that it is consistent with the instability patterns seen in low-precision training.

\subsection{Behavior of the Noise} Let \begin{equation}
    \veps_t \equiv \widetilde{g}_t - \bar{g}_t,
\end{equation}
where $\bar{g}_t$ denotes the exact gradient at time step $t$, and $\widetilde{g}_t$ is its low-precision counterpart. Under a multiplicative noise model, we posit that
\begin{equation}
    \widetilde{g}_t = (1 + \bzeta_t)\bar{g}_t, \label{eq:mult_grad_noise}
\end{equation}
where $\bzeta_t$ is a (possibly data and parameter-dependent) noise matrix induced by quantization. Although $\bzeta_t$ is not directly measurable (and may not even be uniquely defined e.g., due to weight permutations), we can estimate the magnitude of its effect. Specifically, the deviation vector $\veps_t$ satisfies
\begin{equation}
    \|\veps_t\|_2 \le \|\bzeta_t\|_{\textrm{op}} \|\bar{g}_t\|_2, \label{eq:eps_t_bound}
\end{equation}
where $\|\cdot\|_{\textrm{op}}$ denotes the operator norm. In~\Cref{sec:crude_theory_bound}, we argue for a heuristic bound that $\|\bzeta_t\|_{\textrm{op}}$ must satisfy through training and how a runaway loss divergence may occur in this model.

To test this model empirically, we replicate the synthetic experiment setup from \Cref{sec:synthetic}. For each configuration, we fix the random seed and weight initialization, then train one model in FP32 to log the exact gradient $\bar{g}_t$ at each step. We then retrain the same model under MXFP8 precision and compute the deviation $\veps_t = \widetilde{g}_t - \bar{g}_t$ at every step. This allows us to extract both the norm ratio $\|\veps_t\|_2 / \|\bar{g}_t\|_2$ and the cosine similarity between $\widetilde{g}_t$ and $\bar{g}_t$.

\begin{figure}[ht]
  \centering
    \includegraphics[width=\linewidth]{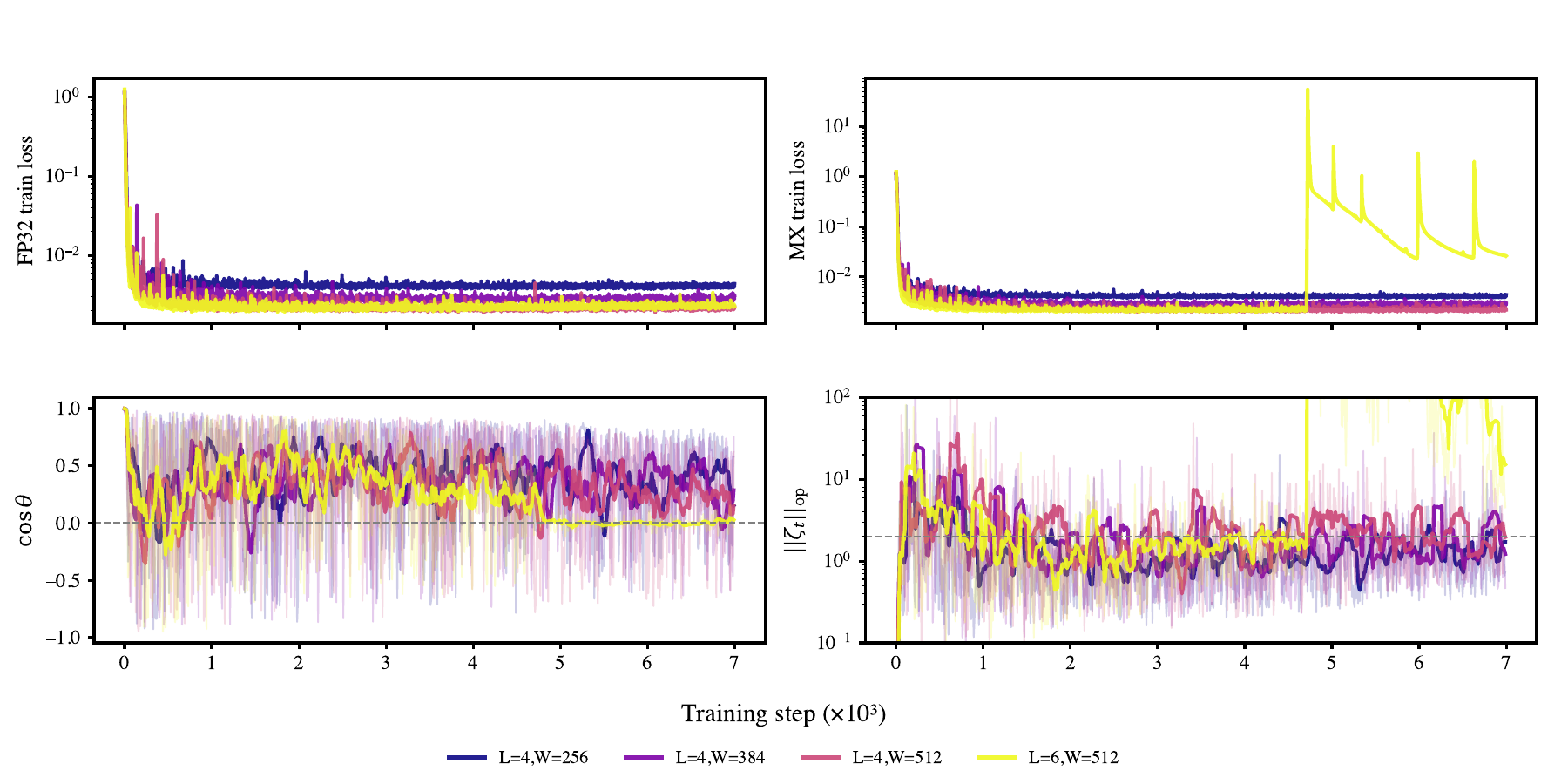} 
  \caption{Shows the bound on the operator norm $\|\bzeta_t\|_{\textrm{op}}$ (as inferred from \Cref{eq:eps_t_bound}), and the cosine angle between the low precision gradient and high precision gradient. Dashed line in the lower right plot shows when the bound on $\|\bzeta_t\|_{\textrm{op}}$ is equal to 2.}
  \label{fig:stability_zeta}
\end{figure}

Results are shown in \Cref{fig:stability_zeta}. Early in training, the estimate of $\|\bzeta_t\|_{\textrm{op}}$ (as inferred from \Cref{eq:eps_t_bound}) gradually decreases. However, as training progresses, the estimate begins to rise. Once $\|\bzeta_t\|_{\textrm{op}} \sim 2$, we observe divergence in the loss. A similar trend is observed in the cosine angle between gradients: it slowly degrades over several thousand steps and eventually flatlines near zero, indicating that the low-precision gradient is no longer aligned with the true descent direction.

\subsection{A Crude Bound\label{sec:crude_theory_bound}} To understand the behavior of $\opnorm{\boldsymbol{\zeta}}$, consider that we have some optimum $w_*$ such that $\nabla_w L(w_*) \approx 0$.  Linearizing around the minimum we have \be \nabla_w L(w_t) = H (w_t - w_*), \ee where $H = \nabla^2_w L$ is the Hessian. The equation above makes no reference to precision -- the only approximation we've made is ignore terms of order $(w_t-w_*)^2$ and higher.  Defining $\delta_t \equiv w_t - w_*$, we then have \be \bar{g}_t = H \delta_t. \ee  With some manipulations the GD update rule is\footnote{Strictly speaking, we are using the stochastic Adam update rule and not GD in our experiments, and so the resulting bound should not be regarded as rigorous.} \be \delta_{t+1} = \delta_t - \eta_t (I + \bzeta_t) H \delta_t \ee and so \be \delta_{t+1} = (I - \eta_t H)\delta_t - \eta_t \bzeta_t H \delta_t. \ee  We can therefore see that there is a driving term  proportional to the noise $\bzeta_t$; if the noise operator norm is large enough, it can flip a contracting direction into an expanding one. The stability criteria is therefore that the operator $I - \eta_t ( 1 + \bzeta_t) H$ has spectral radius less than one.  In terms of the maximum eigenvalue of $H$, $\lambdamax$, this means that a crude bound for stability is \be  |1 - \eta_t \lambda_{\textrm{max}}| + \eta_t \norm{\bzeta_t}_{\textrm{op}} \lambda_{\textrm{max}}  \lesssim 1. \label{eq:simple_bound} \ee  Clearly, when the norm of $\bzeta_t$ grows, the region of stable $\eta_t \lambda_{\textrm{max}}$ shrinks.  However, from the ``edge of stability'' viewpoint of~\citet{DBLP:conf/iclr/CohenKLKT21}, in the absence of multiplicative noise, $\lambda_{\textrm{max}}$ is expected to increase until it hovers at or just above $\sim 2/\eta$. Once the multiplicative term $\bzeta_t$ is introduced, we may then expect that the stability region defined by~\Cref{eq:simple_bound} contracts. Developing a precise theory for this regime -- building on the analysis of~\cite{jastrzebski2020breakevenpointoptimizationtrajectories, damianselfstabilization,DBLP:conf/iclr/CohenKLKT21} -- is an interesting direction for future work.  In the meantime, we bypass an explicit spectral calculation by estimating a lower bound on $\norm{\bzeta_t}_{\textrm{op}}$ directly in our synthetic experiments through~\Cref{eq:eps_t_bound}.  Empirically, we observe a pattern where the running average of this lower bound first drifts downward, later turns upward (lower right of~\Cref{fig:stability_zeta}). When it stabilizes around $\approx2$, training instabilities tend to follow; this observation marks a strong (but not perfect) qualitative correlate of divergence.

\section{What can cause the noise? \label{sec:noise_cause}} Bound~\eqref{eq:simple_bound} predicts roughly \emph{when} instability occurs but not \emph{why} \(\|\boldsymbol\zeta_t\|_{\mathrm{op}}\) grows. Typically, instabilities in low precision happen due to over/underflow or clamping issues that can bias the gradient.  However, in a block scaling format such as MX, it is unclear how such gradient bias may accumulate when the shared scale explicitly puts nearly all values within a representable range. 

\subsection{Overflow Issues with Layernorms\label{sec:LN_issues}}

\begin{figure}[h]
  \centering
    \includegraphics[width=\linewidth]{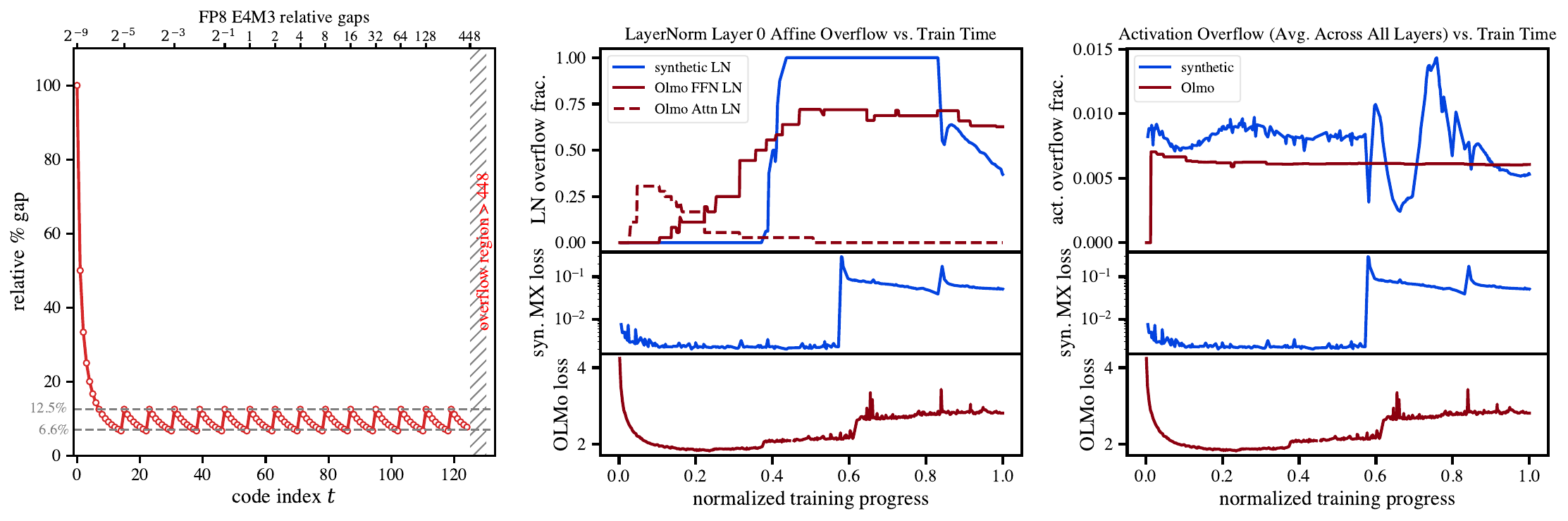} 
  \caption{\textbf{Left}: relative gap $(x_{t+1}\!-\!x_t)/x_t$ for successive
positive FP8 \texttt{E4M3} codes (sign bit stripped).  Within each
exponent band the gap decays from $12.5\%$ to $6.6\%$; the hatched
region marks values that would be clamped once the scaled magnitude
exceeds the representable limit of 448. \textbf{Center}: Top subplot shows what fraction of layernorm affine parameters end up in the last quantization bin after division of the shared scale in the first layer of the network.  For OLMo, we look at the FFN layernorm and the attention layernorm.  The synthetic loss in this case exhibits a divergence in MX precision (but is stable in FP32 precision), and corresponds to the student-teacher setup of~\Cref{eq:student_teacher} with four layers and $d_{\textrm{model}}=512$ and $\eta = 6 \times 10^{-4}$. \textbf{Right}: Shows the fraction of activation values that end up in the last quantization bin after division by the shared scale.  We average across all layers for both synthetic and OLMo runs.}
  \label{fig:ln_act_of}
\end{figure}

To understand this, we begin by examining a concrete example of MXFP8 \verb|E4M3| as specified in~\citet{ocp_mx}.  The left panel of Fig.~\ref{fig:ln_act_of} plots the relative gap \(
(x_{t+1}-x_t)/x_t \) between successive \emph{positive} codes in this format, ordered from index $0$ (the smallest sub-normal, $2^{-9}$) up to index $125$ (448). The index stops at 125 (rather than the expected $2^7-1 = 127$) because S 1111 111$_2$ is reserved for the NaN symbol, which would otherwise correspond to a value of 480, and S 0000 000$_2$ is the zero code, leaving 126 remaining codes~\citep{ocp_mx}.  We can note the following:  

\begin{InsightBox}
\textbf{1.} For a fixed exponent bin the relative gap starts at $12.5\%$ and decays to $6.6\%$ as the mantissa increases.

\textbf{2.} There is an overflow region (left of~\Cref{fig:ln_act_of}) when the value becomes larger than the largest representable normal number (448). Typically, these values are clamped down to 448.
\end{InsightBox}

The latter observation above means that if a block of values lies within a sufficiently small band, these values may end up in the gray overflow region of~\Cref{fig:ln_act_of} after dividing by the block scale.  For example, from~\Cref{algo:convert_to_mx}, for the case of MXFP8 \verb|E4M3| which has $e_{\textrm{max}}^{\textrm{elem}} = 8$ the overflow criteria for a given value $v$ within a block with a shared scale $X$ is \be \left| \frac{v}{X} \right| > 448 \Rightarrow |v| > 0.875 \times (\textrm{abs. max within block}). \label{eq:bound} \ee This type of overflow region was flagged for the case of narrower MXFP4 format in~\citet{tseng2025trainingllmsmxfp4}.  We show that, while MXFP8 E4M3 has a larger dynamic range, the same effect becomes consequential in practice because layernorm \emph{affine} weights are tightly clustered and particularly susceptible to having \textit{all} values within a block falling in this range.  For example, layernorm weights typically follow log-normal distributions with scale $e^\mu \sim 1$ and deviation $ \sigma \ll 1$, and so a block of weights might look something like \begin{InsightBox}
    \verb|[0.89740956, 0.89628334, 0.88358812, 0.88474816, 0.90372837 | \dots \verb|]|
\end{InsightBox} which all end up in the overflow region of~\Cref{fig:ln_act_of} after dividing by $X = 2^{\floor{\log_2(\textrm{abs. max})} - e_{\textrm{max}}^{\textrm{elem}}} = 2^{-8}$.  In our experiments, the impact of this effect is shown in the middle plot of~\Cref{fig:ln_act_of}.  In the synthetic case, in some cases, nearly all of the layer norm weights fall within the band required to flow into the last bucket, losing heterogeneity in nearly all blocks when they are clamped to the maximum normal value after scale division.  Note that this explains, at least partially, why removing the layernorms stabilized low-precision training in~\Cref{fig:act_sweep_without_ln}.  While a different format, like MXFP8 \texttt{E5M2} may avoid this issue, the loss of precision from having only two mantissa bits is a different source of bias and still leads to training instabilities.

In OLMo, there are a large number of layernorms which experience different degrees of clamping to the last quantization bin. Some components, such as the attention layernorms, remain relatively well behaved throughout training, whereas others--like the FFN layernorms or the \( QK \) layernorms~\citep{querykey_ln} can experience large, sudden overflow issues.  While it's possible to disable the affine transformation of layernorms, this greatly diminishes performance in the language model setting.  More broadly, this issue indicates a problem with applying shared-scales to blocks of weights that follow approximately log-normal distributions (such as layernorm affine parameters), which may not have a well-defined notion of a ``max" relative to a resolution fixed by a given precision scheme.  A scale that adapts to both \emph{min} and \emph{max} might avoid the bias; we defer this to future work.  On the activation side, we find that this effect does affect roughly $\sim$1\% of values in our synthetic experiments and $\sim$0.5\% of values in OLMo (shown in the right subplot of~\Cref{fig:ln_act_of}). 

\subsection{Potential Mitigations}
Given these observations, we next ask whether training instabilities can be mitigated by modifying the quantization scheme, for example by increasing the precision of activation/LN elements or by restricting quantization to the forward pass only. To investigate this, we perform a sweep over model sizes comparing three setups: (1) full quantization in both forward and backward passes (baseline), (2) forward-pass-only quantization, and (3) higher-precision activation formats (including layernorms) in the MX scheme.

\begin{figure}[ht]
  \centering
    \centering
    \includegraphics[width=\textwidth]{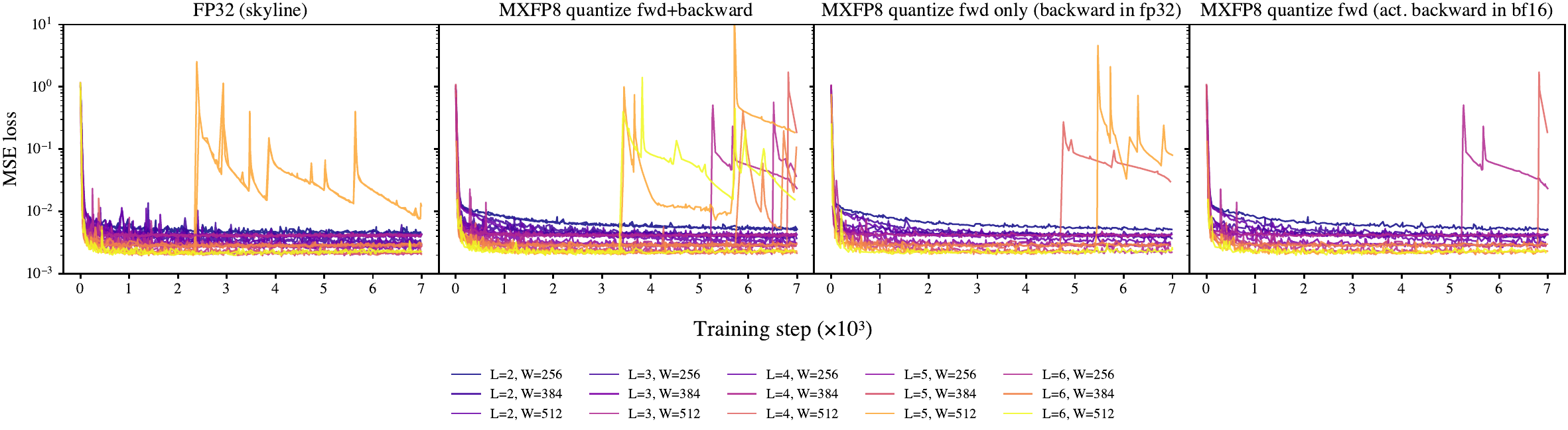} 

  \hfill
  \caption{Shows two mitigations (quantization of only forward pass) and activation elements in higher precision, compared to fully quantized baseline and FP32 skyline.}
  \label{fig:synthetic_mitigations}
\end{figure}

As shown in~\Cref{fig:synthetic_mitigations}, both mitigation strategies do improve stability. Each reduces the number of divergent runs to 2, down from 6 in the fully quantized baseline. One of the instability observed in the forward-only setting also occurs in high-precision training, leaving only one instability unique to the MXFP8 \texttt{E4M3} baseline unmitigated.

\begin{figure}[h]
  \centering
    \includegraphics[width=\linewidth]{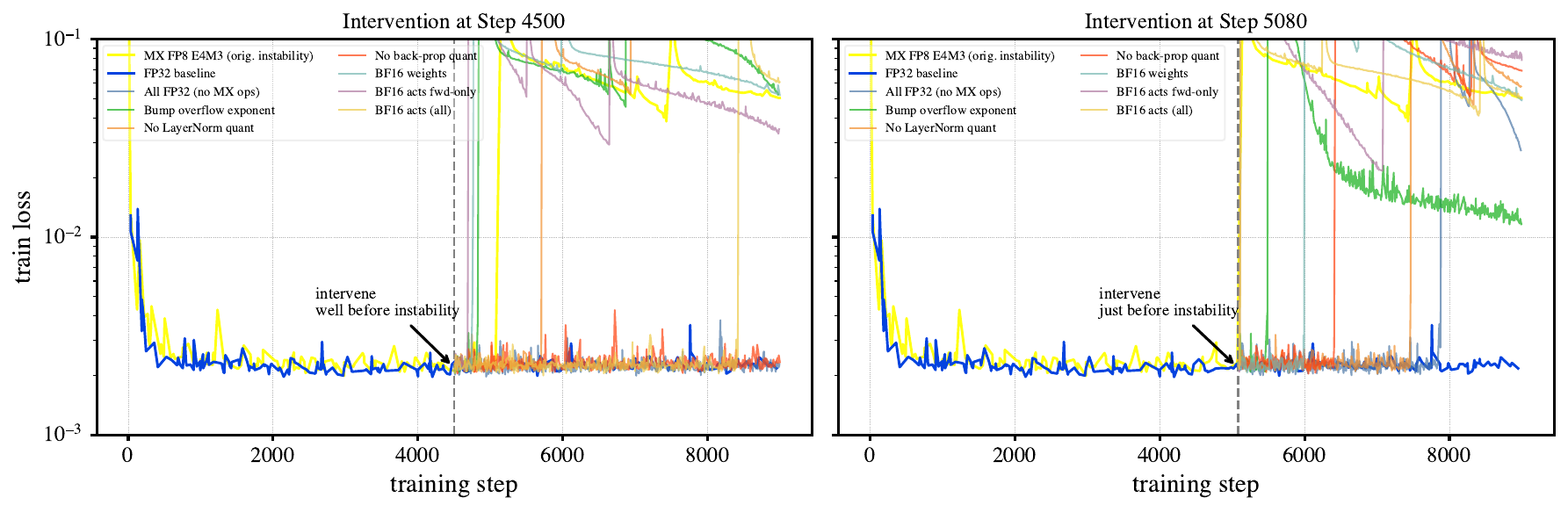} 
    \caption{Intervention experiment for a synthetic student-teacher model with $d_{\textrm{model}}{=}512$, four layers, and learning rate $\eta{=}6{\times}10^{-4}$. Training is stable in FP32 (blue) but diverges in MXFP8 \texttt{E4M3} (yellow) around step 5100. We test two intervention timings: step 4500 (left, well before instability) and step 5080 (right, just before instability). Early interventions, like disabling backward-pass quantization or switching to high-precision (FP32), successfully prevent divergence, while using high precision for the activations (bfloat16) can greatly delay it. Late interventions cannot avert instability but can only delay it; the most effective are switching to FP32 or skipping quantization of layernorm weights.}
  \label{fig:synthetic_interventions}
\end{figure}

We can actually go further and ask whether an impending divergence can be averted by \emph{in-situ} interventions to the training recipe.  Figure~\ref{fig:synthetic_interventions} tracks a configuration that is stable in FP32 but diverges in MXFP8 \texttt{E4M3}. This setting corresponds to the previously described student-teacher scenario with four layers and model dimension $d_{\textrm{model}} = 512$. The instability starts approximately at step 5090 and we consider interventions just before the instability at step 5080 and well before the instability at step 4500. For each intervention we keep the random seed, model state, and batch sequence identical, so the training state at the intervention step is the same as in the baseline run; any divergence afterward is therefore attributable to the intervention.

\begin{itemize}
\item \textbf{Switching entirely to FP32 precision for remaining training steps.} Implementing FP32 significantly stabilizes training if the change is made sufficiently early (step 4500), but it is ineffective if applied immediately before instability (step 5080). However, even at the later intervention, FP32 prolongs training stability more effectively than the other approaches.

\item \textbf{Increasing the shared exponent by one (bumping exponent).} Adjusting the exponent to avoid the last bucket overflow for blocks that have values that fall into the range in~\Cref{eq:bound} does not  mitigate instability, possibly due to insufficient precision improvement from a single increment.

\item \textbf{Avoiding MX quantization for LayerNorm affine parameters.} Omitting quantization of these parameters partially stabilizes training and delays instability significantly at both intervention steps, indicating that LayerNorm parameters do contribute to instability dynamics. However, eventual instability suggests a residual effect from quantized activations.

\item \textbf{Precision adjustments in forward and backward passes,} where we explored:
    \begin{itemize}
        \item quantizing weights and activations only during the forward pass (no backward-pass quantization);
        \item maintaining weights in bfloat16 and activations in MXFP8 (both passes);
        \item maintaining activations in bfloat16 for the forward pass but MXFP8 for backward (with MXFP8 weights);
        \item using BF16 activations for both forward and backward passes while quantizing weights with MXFP8.
    \end{itemize}
    Among these, applying the intervention just before instability (step 5080), bfloat16 activation precision in both passes consistently provides the strongest immediate stabilization, closely followed by disabling backward-pass quantization. When interventions occur earlier (step 4500), not quantizing the backward step performs comparably to the FP32 baseline, while fully bfloat16 activations delay instability considerably yet eventually become unstable. These results are consistent with a stochastic noise model, suggesting that multiple interacting factors influence instability likelihood. They are also consistent with our earlier observations about the effectiveness of higher activation precision and forward-only quantization schemes in mitigating instability in MXFP8 training.  The mixed weight/activation precision strategy may be a pragmatic approach, as long as careful attention is given to the behavior of layernorm weights throughout training.
\end{itemize}

\paragraph{Other Sweeps} In Appendix~\ref{app:additional_syn_sweeps}, we report additional ablations over optimizer choices (SGD with and without momentum, and Adam) and weight initialization schemes with reduced variance. While these variations can partially reduce the frequency of instabilities, they do not address the underlying bias in the computation. As we show in~\Cref{sec:theory} and in~\Cref{sec:LN_issues}, many instabilities stem from biased gradient estimates, and these choices do not reliably eliminate that source of error.

\paragraph{Key Takewaways} Our synthetic experiments reveal that training instabilities in block-scaled  low-precision settings arise from both stochastic optimization effects and quantization-induced bias. These failures result from a complex and subtle interplay between architectural choices, activation quantization, quantization of layer normalization parameters, and learning rate. The dominant precision-specific bias comes from overflow of tightly clustered layer-norm affine weights (and a small fraction of activations).  Our intervention experiments show that while simple tweaks such as nudging the shared exponent do not remove this bias, mitigations which raise precision in key parts of the computation -- such as increasing the precision of activations or not quantizing in the backward step -- generally improve stability.

\section{Stabilization Strategies in Language Model Setting \label{sec:llm_scaling}} Motivated by the effective mitigations observed in our synthetic experiments, we return to the language model (OLMo) setting and apply the same two strategies: (1) retaining \texttt{bfloat16} as the element format for activations and layer-norms, and (2) applying MX quantization only to the forward pass. In both cases, we find that training remains stable across all FP8 configurations, including \texttt{E4M3} and \texttt{E5M2} formats.

\begin{table}[h]
\centering
\scriptsize
\setlength{\tabcolsep}{5.2pt}
\begin{tabular}{l c c c c c c c c}
\toprule
\textbf{Weight} & \textbf{Activation} & \multicolumn{7}{c}{\textbf{\(D/N\) Ratio}} \\
\cmidrule(lr){3-9}
 &  & \textbf{140.96} & \textbf{99.19} & \textbf{70.91} & \textbf{37.86} & \textbf{21.28} & \textbf{16.23} & \textbf{12.51} \\
  &  & $N$=0.16B & $N$=0.19B & $N$=0.23B & \textbf{}$N$=0.31B & \textbf{}$N$=0.42B & \textbf{}$N$=0.48B & \textbf{}$N$=0.54B \\
\midrule
\rowcolor{gray!10}
bfloat16   & bfloat16   & 0.710 & 0.703 & 0.698 & 0.691 & 0.688 & 0.686 & 0.686 \\
\midrule
\texttt{MXFP8 E4M3} & bfloat16   & 0.0 & -0.002 & -0.002 & 0.0 &  0.0 & 0.0 & 0.0 \\
\texttt{MXFP8 E5M2} & bfloat16   & 0.105 & 0.107 & 0.112 & 0.004 & 0.002 & -0.001 & -0.001  \\
\midrule
\texttt{MXFP8 E4M3} & \texttt{MXFP8 E4M3} & 0.005 & 0.002 & 0.002 & 0.004 & 0.002 & -0.001 & -0.001 \\
\texttt{MXFP8 E5M2} & \texttt{MXFP8 E5M2} & 0.010 & 0.012 & 0.057 & 0.019 & 0.007 & 0.004 & 0.004 \\
\bottomrule \\
\end{tabular}
\caption{The validation loss on Fineweb-Edu of high precision runs versus low precision with mitigations applied (values are shown as differences with respect to bfloat16 baseline; lower is better). For the last two rows, we quantize only the forward pass.} \label{tab:val_loss_compare_after_fix}
\end{table}

Table~\ref{tab:val_loss_compare_after_fix} reports validation loss differences relative to full-bfloat16 baselines. MXFP8 \texttt{E4M3} weights paired with bfloat16 activations in particular match full-precision performance across all tested model sizes.As a proof of concept, Figure~\ref{fig: scaling-laws-high-act} shows a Chinchilla-style scaling law fit to the stabilized MXFP8 \texttt{E4M3} + {bfloat16 runs. This demonstrates that, at least at the scales we trained, valid empirical scaling laws can still be extracted under this hybrid format. Whether this continues to hold at larger scales remains an open question, as such lower precision may exhibit bottoming-out effects or other nonlinearities in training dynamics. Full loss curves and scaling law fits for both mitigation strategies compared to bfloat16 baselines are provided in~\Cref{sec:lang_model_scaling_and_training_after_fix}.

\section{Conclusion}

We have shown that training large language models in block-scaled low-precision formats (MXFP8, MXFP6) often leads to sharp, unrecoverable instabilities. By combining large-scale LLM sweeps with a controlled student-teacher proxy model trained on synthetic data, we isolate two distinct failure modes: \begin{itemize} \item Stochastic optimization breakdown, driven by a complex interplay between hyperparameters choices (e.g.\ high learning rates) or architectural factors (such as width, depth, choice of activation function), can alone can trigger instabilities even in high precision.
\item Quantization-induced gradient bias, where shared-scale clamping (particularly of layer-norm affine weights and to a lesser extent, other activations) injects multiplicative gradient noise that ultimately destabilizes training.
\end{itemize}

Guided by these observations, we identify two effective mitigation strategies: maintaining higher precision for activations or limiting quantization exclusively to the forward pass. Notably, we find that using MXFP8 \texttt{E4M3} weights in combination with bfloat16 activations matches the performance of full-bfloat16 baselines.  While our implementation applied higher-precision activations network-wide, it is likely that stability could be preserved using higher precision selectively in key layers or modules.

Looking ahead, continued hardware advances will expand the frontier of what’s computationally feasible. Some concrete directions for future research include:
\begin{itemize} \item extending our proxy model to include attention mechanisms, mixture-of-experts with many layers, and other transformer-specific components to better predict instabilities in state-of-the-art architectures;
\item developing a clear theoretical picture of the stochastic noise model and how different factors can amplify or reduce the risk of a training instability;
\item designing new blockwise scaling schemes that adapt to skewed or tightly clustered distributions.
\end{itemize}

\begin{figure}[ht]
  \centering
  \begin{subfigure}[t]{0.44\textwidth}
    \centering
    \includegraphics[width=\linewidth]{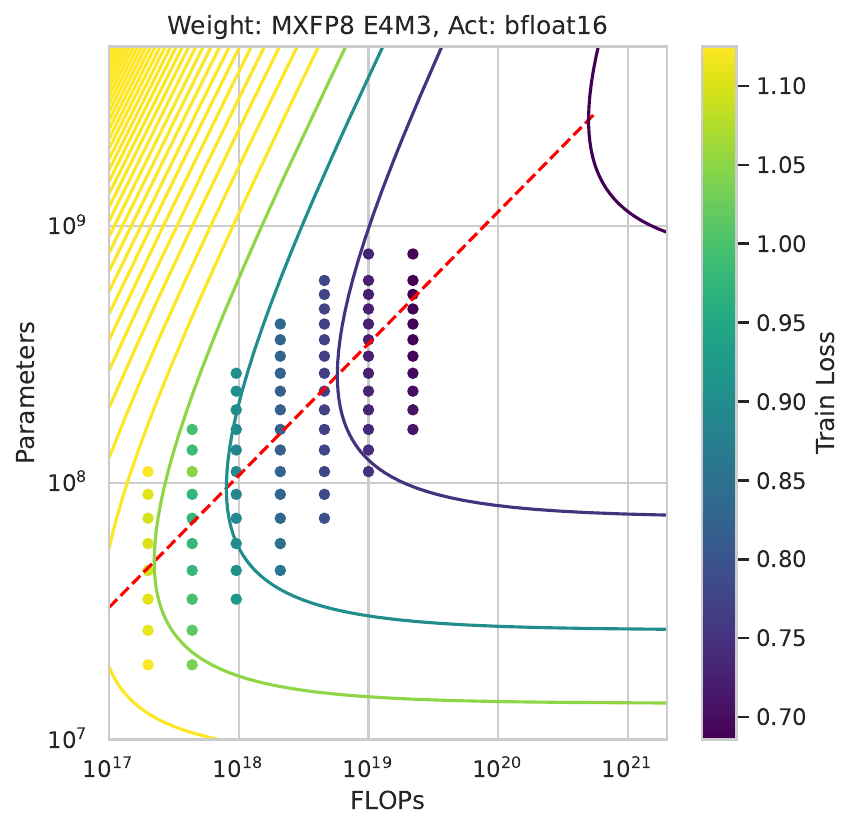} 
    \caption{Scaling law fit for \texttt{FP8 E4M3}-bfloat16.}
    \label{fig: }
  \end{subfigure}
  \hfill
  \begin{subfigure}[t]{0.44\textwidth}
    \centering
    \includegraphics[width=\linewidth]{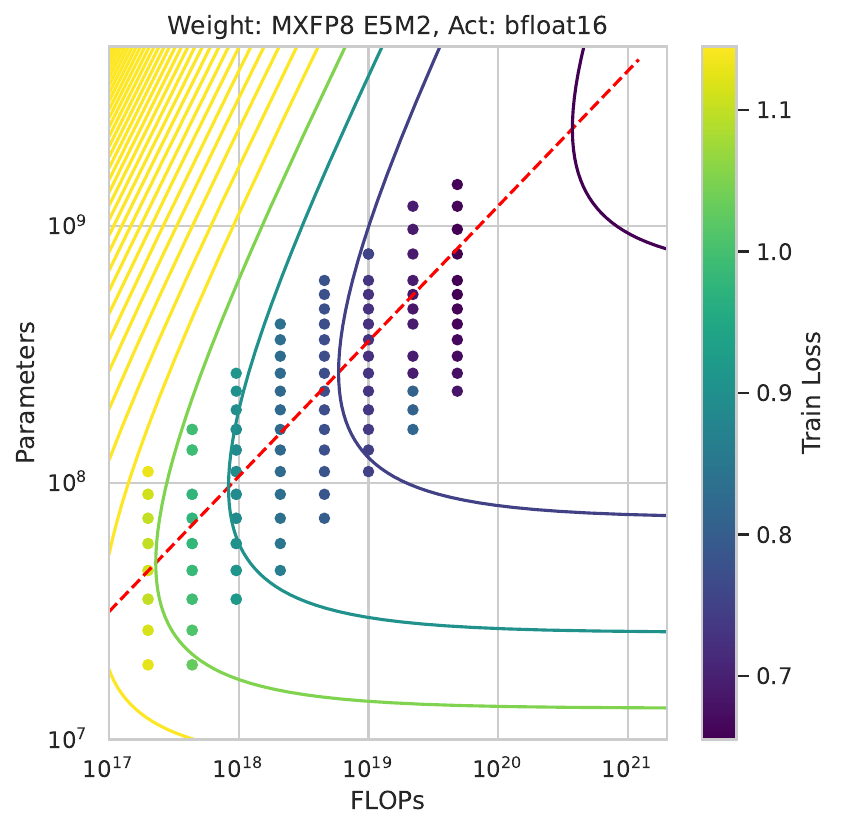} 
    \caption{Scaling law fit for \texttt{FP8 E5M2}-bfloat16.}
    \label{fig: }
  \end{subfigure}
  \caption{Scaling law fit for combinations of precision formats of weights and keep the activations in high precision. Fit was calculated using a Chinchilla model for the loss; details and fit parameters are given in~\Cref{sec:lang_model_scaling_and_training_after_fix}.}
  \label{fig: scaling-laws-high-act}
\end{figure}

\clearpage

\textbf{Limitations}

In this work, we provide insights into when instabilities arise in student-teacher training setups using a residual MLP, and show that these findings generalize to more complex architectures. Specifically, we identify sources of training instability—such as batch size, batch quality, and learning rate—that affect both high- and low-precision regimes. We also highlight instability factors specific to low-precision training, including sensitivity to learning rate, model depth and width, and the choice of activation function (though not exhaustively explored, they represent the primary contributors). To mitigate these instabilities in full-scale models, we propose retaining gradient computations and/or activation functions in full precision. These approaches stabilize training and enable us to fit precision-aware scaling laws, specifically capturing how validation performance scales with model size and number of training tokens.

However, there are many interactions in model training and we cannot exhaustively cover them all in our sweeps. It is likely that there exist other mitigation strategies not evaluated in this work. We emphasize, however, that a core contribution of this paper is to propose simplified proxies for reasoning about complex low-precision dynamics, rather than prescribe a single universal fix. Our results are primarily derived from decoder-only language models and residual MLPs; future work is needed to determine whether similar behaviors arise in MoE architectures. Finally, while we emulate MX formats in PyTorch, all experiments are conducted in software. Real-world deployment on Blackwell-class hardware may introduce additional sources of error due to rounding behavior, memory layout, or fused kernel execution not captured by our implementation.

We also performed most of our experiments on relatively small models; our synthetic models run on NVIDIA H100s in a matter of minutes, and the largest language models used $16$ H100s per run.  While this allowed us to reach the ~1B scale, results may not extrapolate well to models that are an order of magnitude larger.

\clearpage
\appendix

\section{Review of Shared-Scale Quantization\label{app:review_mx}} In this section we provide a self-contained review of block scaling quantization schemes, largely following~\cite{rouhani2023microscaling, ocp_mx}.  Taking a step back, the idea in shared-scale quantization methods is to introduce a number which represents the shared scale among a group of values that could, e.g., represent weights or activations.  The idea is that low-precision data types tend to have a small representable range and quantization can clip very large values or zero-out smaller values.  By dividing by the shared scale, the goal is to put these numbers in a representable range and save the scale such that it may be multiplied at the end of the computation.  There are many choices for how to pick the scale, with pros and cons for each.  For example, one approach is to have a single scale factor for the entire tensor, which has a very low memory overhead but is usually too coarse-grained and lead to saturation issues.  On the opposite end, one could keep a scale factor for every value in the tensor which obviously allows for higher accuracy but involves much more memory.  Other approaches include tilewise scaling, where a scale factor is used for a fixed-size submatrix.  This was the approach taken in~\citet{liu2024deepseek}.  In this work, we focus on \textit{block} scaling methods, where a single 1-dimensional block of values shares a scale.  In particular, we focus on the ``microscaling" (MX) format, where each block consists of 32 values, with a shared scale that can be computed using~\Cref{algo:convert_to_mx}.  When performing matrix multiplications or dot products, these shared scales are carried around and multiplied at the end of the computation (see~\citet{ocp_mx} for the exact specifications).

\begin{algorithm}
\caption{Convert \(\mathbf{V}\in\texttt{HP\_DTYPE}^{k}\) to an MX block \(\{X,\;P\in\texttt{LP\_DTYPE}^{k}\}\)}
\begin{algorithmic}[1]
\Require $k=32$ (hardware block size),\\
         $e_{\text{max}}^{\text{elem}}$ — exponent of the largest normal value in \texttt{LP\_DTYPE}
\Ensure  Scale factor \(X\) and low-precision elements \(P_1,\dots,P_k\)  

\State $m \gets \max_{i}\bigl(\lvert V_i\rvert\bigr)$
\State $\textit{shared\_exp} \gets \bigl\lfloor\log_2(m)\bigr\rfloor - e_{\text{max}}^{\text{elem}}$ \label{ln:shared-exp}
\State $X \gets 2^{\textit{shared\_exp}}$   \Comment{block scale (a power of two)}
\For{$i \gets 1$ \textbf{to} $k$}
    \State $r \gets V_i / X$
    \State $P_i \gets \textsc{QuantizeToLP}(r)$  \Comment{clamp if $|r|$ overflows}
\EndFor
\State \Return $(X,\; \{P_i\}_{i=1}^k)$
\end{algorithmic} \label{algo:convert_to_mx}
\end{algorithm}

The shared scale in MX formats can therefore be regarded as the largest power-of-two that can represent the maximum within a block, shifted by the exponent of the largest normal value in that type. What gets quantized in a typical training setup?  There are several toggles.  Let us illustrate by a simple example i.e. our synthetic model in~\Cref{eq:student_teacher} when there is only one layer: \be \begin{aligned} A_0 &= x \quad\quad\quad h = \bW^{(1)} \LN(x) \\
A_1 &= x + \bW^{(2)} \phi(h)  \end{aligned}. \ee

Roughly speaking, we can choose to apply MX quantization to weights, activations in the forward and/or backward pass.  In the forward pass, we can apply~\Cref{algo:convert_to_mx} at every stage.  If $\widetilde{\cdot}$ denotes the quantization resulting from~\Cref{algo:convert_to_mx} and perform matmuls using the shared scale, then if we quantize both weights and activations we compute \be \begin{aligned} A_0 &= x \quad\quad\quad h = \widetilde{\bW}^{(1)} \LN(\widetilde{x})  \\
A_1 &= x + \widetilde{\bW}^{(2)}  \phi(\widetilde{h}).  \end{aligned} \ee  
One additional subtlety is that vector operations such as vector addition such as those present in layernorm computations are typically carried out in bfloat16, \textit{i.e.} the operands are first cast once to bfloat16, and the addition itself runs in bfloat16. Meanwhile, in the backward pass, we can actually quantize in three different ways (or not at all if we turn off quantization in backpropagation): quantize weights, quantize input activation gradients, or quantize output activation gradients.  Suppose we want to compute $\nabla_{\bW^{(1)}} L$ with the MSE loss $ L = \frac{1}{2}(A_1 - y_*)^2$.  This means computing $\pdr{L}{A_1}\pdr{A_1}{\bW^{(1)}}$.  The term $\pdr{L}{A_1}$ can be quantized in one format -- the output gradient quantization, while the second term $\pdr{A_1}{\bW^{(1)}}$ can be quantized in a separate format -- the input gradient quantization.  In general, in this work, unless we state otherwise, these two formats are taken to be the same.  In backpropagating, the input gradient itself will involve computing $\pdr{\phi}{h}$. Again, the exterior gradient $\pdr{\phi}{h} \pdr{h}{W^{(1)}}$ is quantized according to its chosen format.  The second piece, $\pdr{h}{\bW^{(1)}}$ involves the input to the linear layer $\LN(x)$ which is quantized to the input gradient format.

\section{Additional Synthetic Sweeps \label{app:additional_syn_sweeps}} 

In this section, we present additional synthetic experiments to further examine the sources and mitigation of low-precision instabilities.

\Cref{fig:spike_counts} summarizes the frequency of instability spikes across our depth-width sweep at a fixed learning rate of $\eta = 5 \times 10^{-4}$. The MX-mix format refers to the asymmetric configuration using \texttt{MXFP8 E4M3} in the forward pass and \texttt{E5M2} in the backward pass.  Spikes were determined by the heuristic criteria that the loss at time step $t$ had to be a factor of 100 lager than the loss at time step $t-1$; this gives a rough lower bound on the number of spikes.

\Cref{fig:sgd_ablation} compares the impact of optimizer choice, focusing on SGD with momentum, and vanilla SGD (momentum = 0). These experiments used a slightly higher learning rate of $\eta = 1 \times 10^{-2}$ to exaggerate differences. Compared with~\Cref{fig:synthetic_LR_sweep}, we observe that SGD variants are more stable than Adam, perhaps due to Adam's use of second-moment accumulation, which may amplify quantization-induced bias in low-precision regimes.

\Cref{fig:weight_init_ablation} evaluates the effect of different weight initialization schemes. We compare standard Pytorch initialization, typically taken to be a Kaiming uniform distribution between $[-1/\sqrt{\textrm{fan in}}, 1/\sqrt{\textrm{fan in}}]$, against a variant using lower gain ($\texttt{gain} = 0.5$) under the Xavier normal distribution. Reducing the variance of initial weights appears to improve loss spikes.

\begin{figure}[ht]
    \centering
    \includegraphics[width=.95\linewidth]{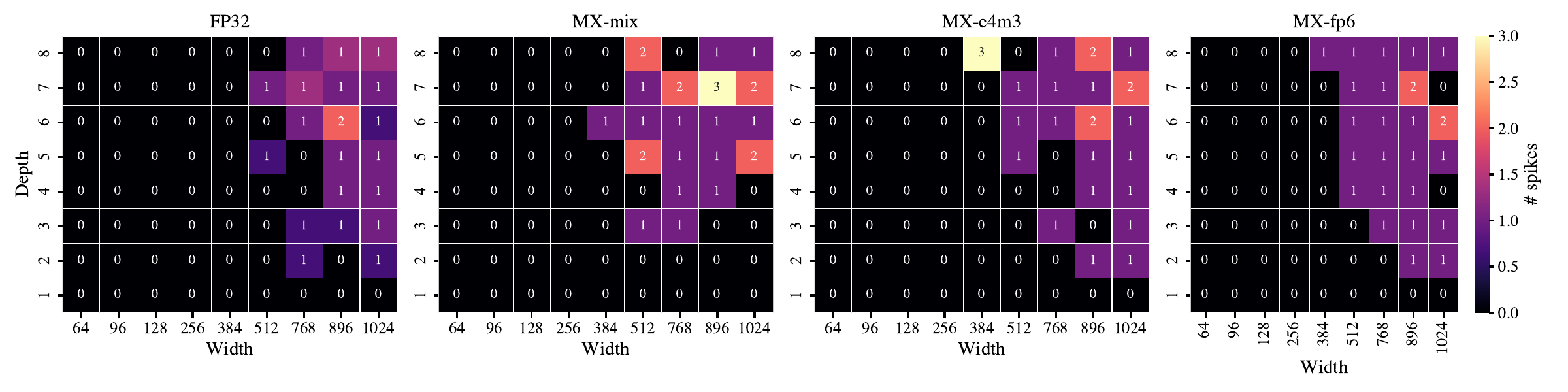} 
    \label{fig: }
  \hfill
  \caption{Instability spikes measured in training, for different model depths and widths.}
  \label{fig:spike_counts}
\end{figure}

\begin{figure}[ht]
  \centering
    \includegraphics[width=\linewidth]{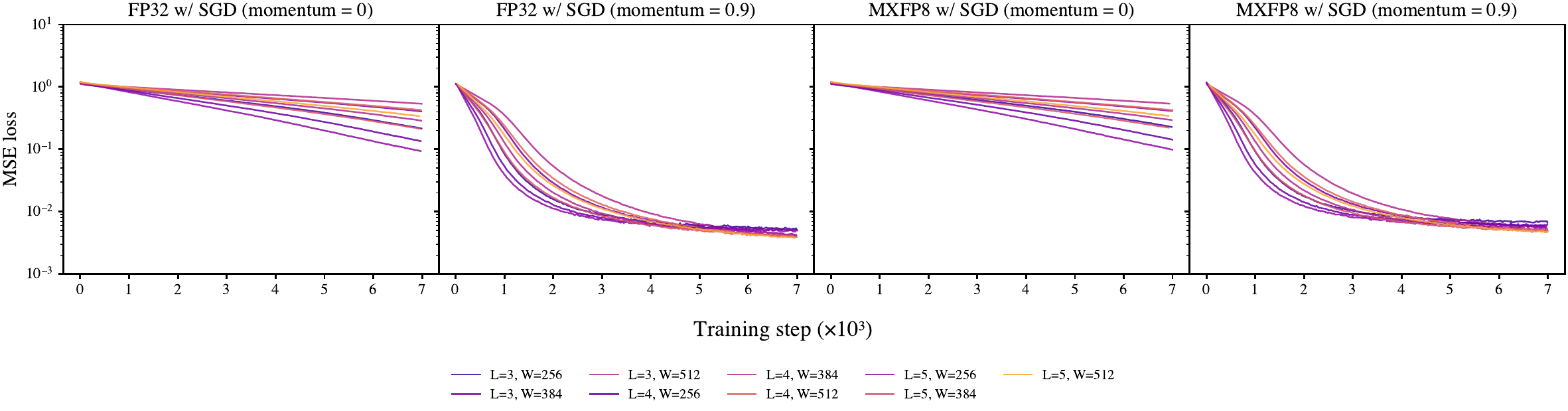} 
    \label{fig: }
  \caption{SGD with and without momentum; a larger learning rate was used $\eta = 1\times 10^{-2}$.}
  \label{fig:sgd_ablation}
\end{figure}

\begin{figure}[ht]
  \centering
    \includegraphics[width=\linewidth]{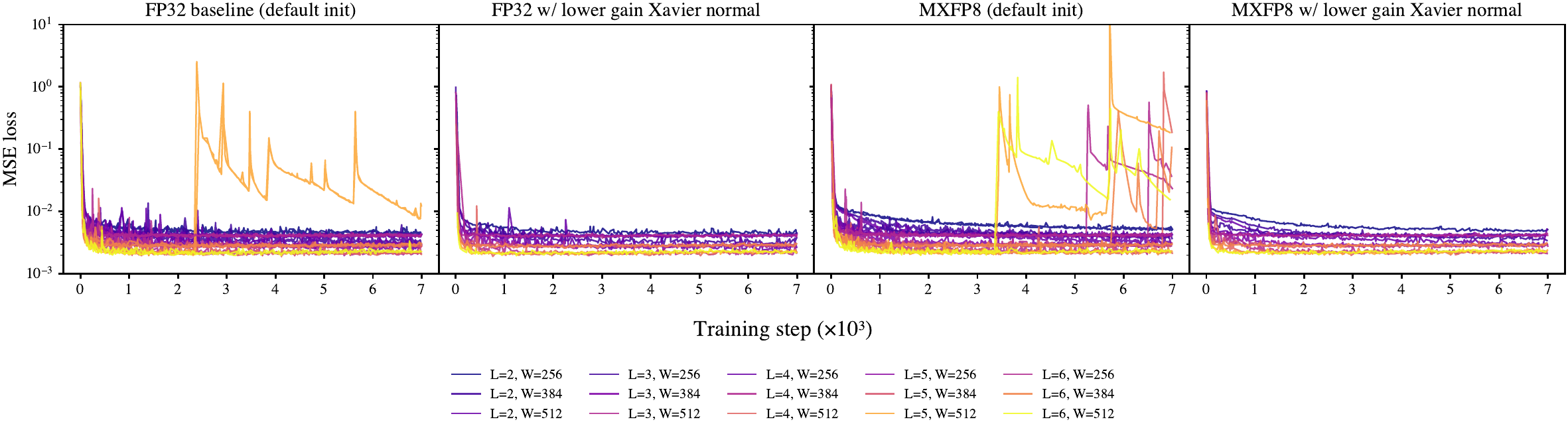} 
    \label{fig: }
  \caption{Baseline versus using a lower gain Xavier normal weight initialization.}
  \label{fig:weight_init_ablation}
\end{figure}

\clearpage
\section{Scaling Law Fits and Loss Curves after Mitigation\label{sec:lang_model_scaling_and_training_after_fix}} 
In addition to~\Cref{fig: scaling-laws-high-act} we provide scaling law for the mitigation where we quantize only the forward pass; this is shown in~\Cref{fig: scaling-laws-fwd-only} which can be compared against the bfloat16 baseline in~\Cref{fig: bfloat16_scaling_baseline}.  Scaling law fits were performed using the methods described in~\cite{hoffmann2022training, brandfonbrener2024loss} where the validation loss was fit with a functional form \be L(N, D) = E + \frac{A}{N^{\alpha}} + \frac{B}{D^{\beta}}, \ee for constants $A$, $B$, $E$, $\alpha$, and $\beta$.  The fitted values of these constants are given in~\Cref{tab:fitted_scaling_law_params}.

We also provide the loss curves after implementing these mitigation strategies; these are shown in~\Cref{fig: stable-quantize-fwd-only} and~\Cref{fig: stable-high-activations}.

\begin{figure}[ht]
  \centering
  \begin{subfigure}[t]{0.49\textwidth}
    \centering
    \includegraphics[width=\linewidth]{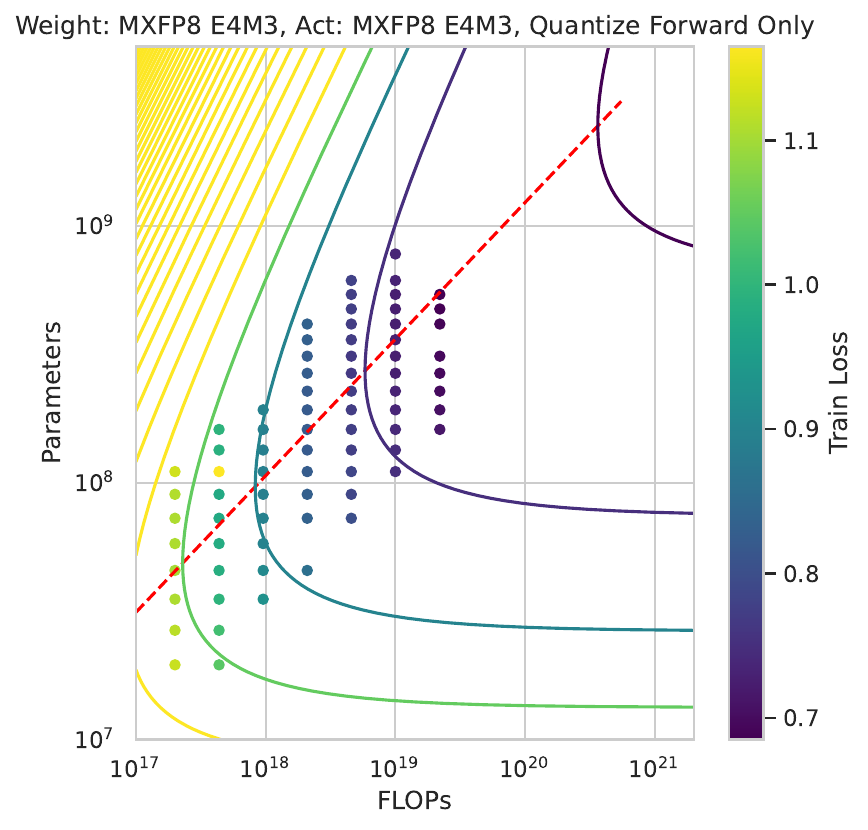} 
    \caption{Scaling law fit for MXFP8-\texttt{FP8 E4M3}.}
    \label{fig: }
  \end{subfigure}
  \hfill
  \begin{subfigure}[t]{0.49\textwidth}
    \centering
    \includegraphics[width=\linewidth]{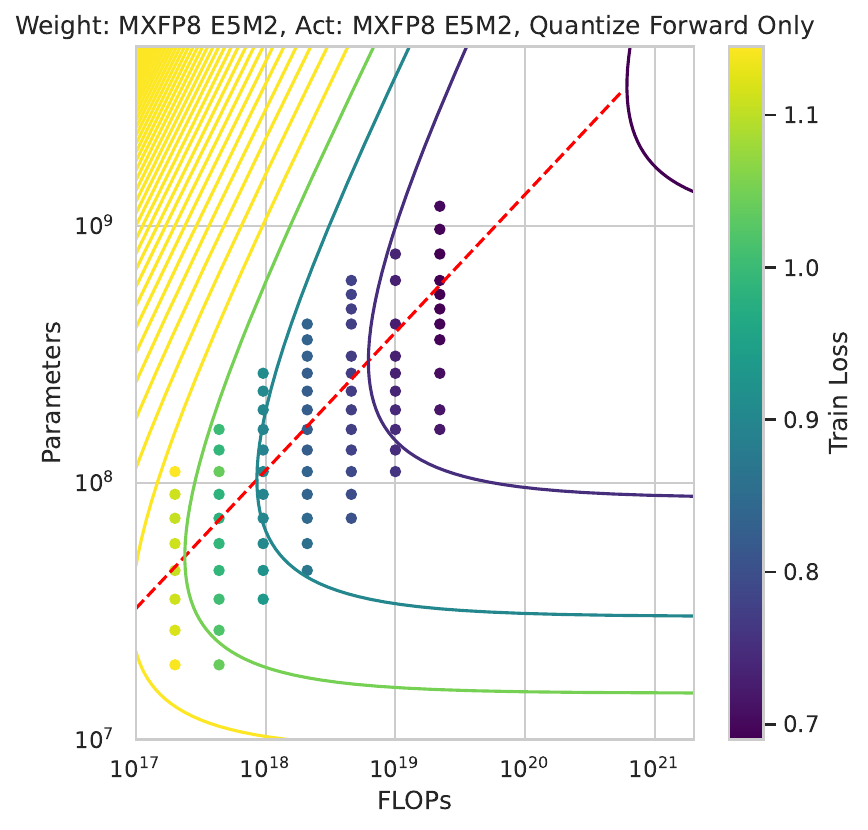} 
    \caption{Scaling law fit for \texttt{MXFP8 E5M2}-\texttt{E5M2}.}
    \label{fig: }
  \end{subfigure}
  \caption{Scaling law fits for fixed stable of precision formats of weights and activations quantizing only the forward pass.}
  \label{fig: scaling-laws-fwd-only}
\end{figure}

\begin{figure}[ht]
  \centering
  \begin{subfigure}[t]{0.49\textwidth}
    \centering
    \includegraphics[width=\linewidth]{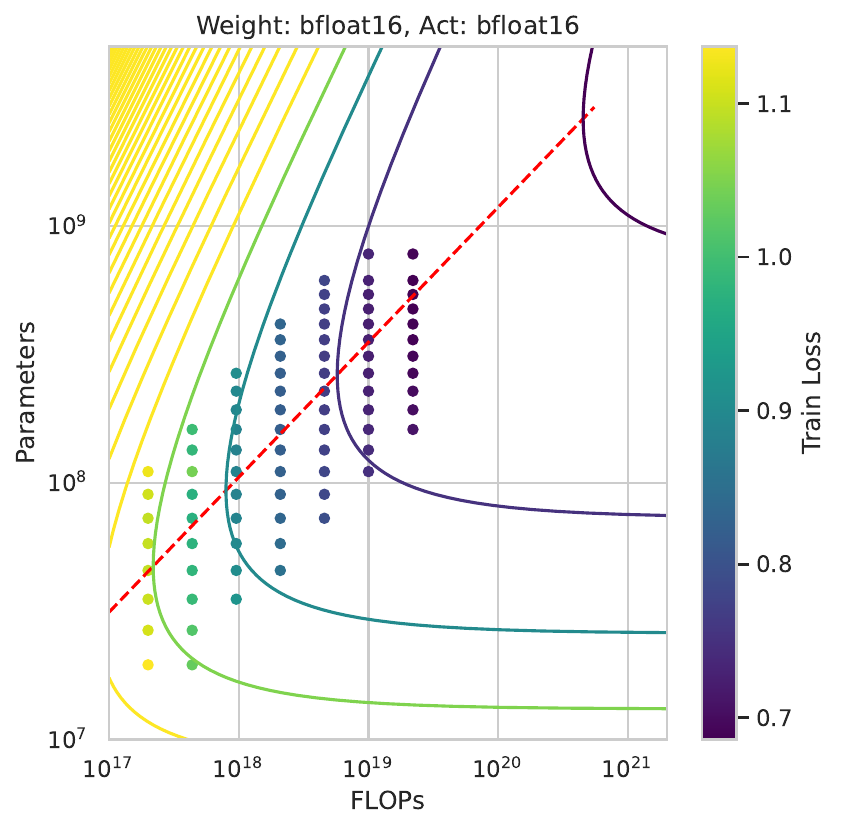} 
    \caption{Scaling law fit for bfloat16-bfloat16.}
    \label{fig: }
  \end{subfigure}
  \hfill
  \begin{subfigure}[t]{0.49\textwidth}
    \centering
    \includegraphics[width=\linewidth]{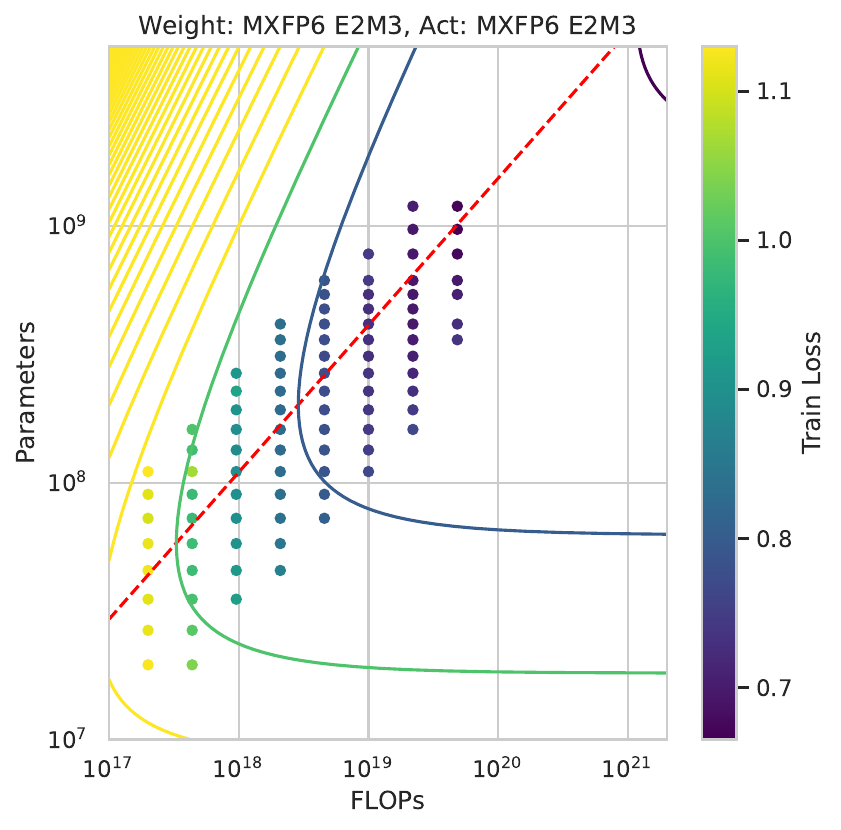} 
    \caption{Scaling law fit for \texttt{FP6 E2M3}-\texttt{FP6 E2M3}.}
  \end{subfigure}
  
  \caption{Scaling law fits for bfloat16-bfloat16 (baseline) and for MXFP6 format.}
  \label{fig: bfloat16_scaling_baseline}
\end{figure}

\begin{table}[h]
\centering
\begin{tabular}{l c c c c c c c c}
\toprule
\textbf{Weight} & \textbf{Activation} & \textbf{A} & \textbf{B} & \textbf{E} & $\boldsymbol{\alpha}$ & $\boldsymbol{\beta}$ & $\boldsymbol{a}$ \\
\midrule
\texttt{MXFP6 E2M3} & bfloat16 & 1.84e+03 & 8.77e+03 & 0.52 & 0.50 & 0.51 & 0.51\\
\texttt{MXFP8 E4M3} & bfloat16 & 2.82e+03 & 2.04e+04 & 0.54 & 0.52 & 0.55 & 0.51 \\
\texttt{MXFP8 E5M2} & bfloat16 & 1.68e+03 & 1.84e+04 & 0.52 & 0.49 & 0.55 & 0.53 \\
bfloat16 & bfloat16 & 1.94e+03 & 2.18e+04 & 0.53 & 0.50 & 0.56 & 0.53 \\
\midrule
\texttt{MXFP8 E4M3} & \texttt{MXFP8 E4M3} & 1.57e+03 & 2.11e+04 & 0.52 & 0.49 & 0.55 & 0.53 \\
\texttt{MXFP8 E5M2} & \texttt{MXFP8 E5M2} & 2.20e+03 & 3.98e+04 & 0.54 & 0.51 & 0.59 & 0.54 \\
\bottomrule \\
\end{tabular}

\caption{Fitted scaling law parameters. For the last two rows, we quantize only the forward pass. The last column is equal to the ratio $a=\beta/(\alpha+\beta)$, the exponent of the optimal model size relative to FLOPs.}
\label{tab:fitted_scaling_law_params}
\end{table}

\begin{figure}[ht]
  \centering
  \begin{subfigure}[t]{\textwidth}
    \centering
    \includegraphics[width=\linewidth]{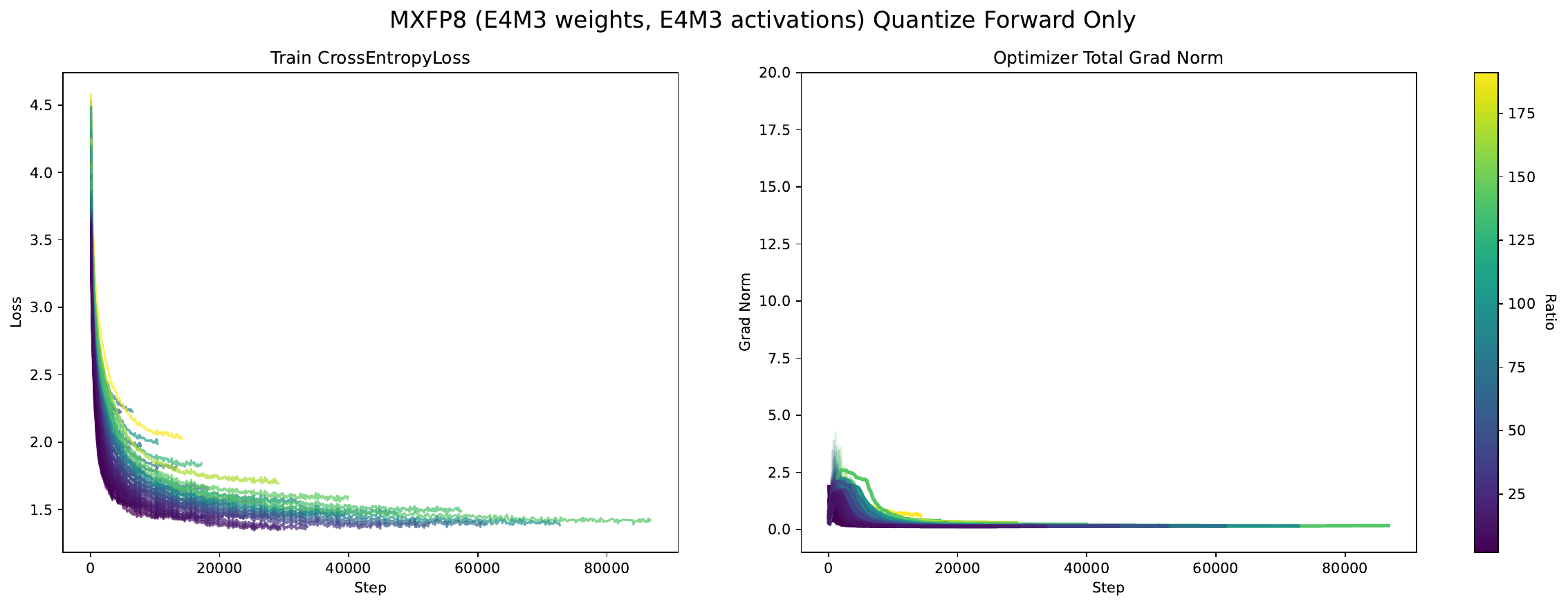} 
    \caption{Train loss and gradient norm for weights: \texttt{MXFP8 E4M3}, Activation: \texttt{MXFP8 E4M3} while quantizing only the forward pass.}
    \label{fig: }
  \end{subfigure}
  \hfill
  \begin{subfigure}[t]{\textwidth}
    \centering
    \includegraphics[width=\linewidth]{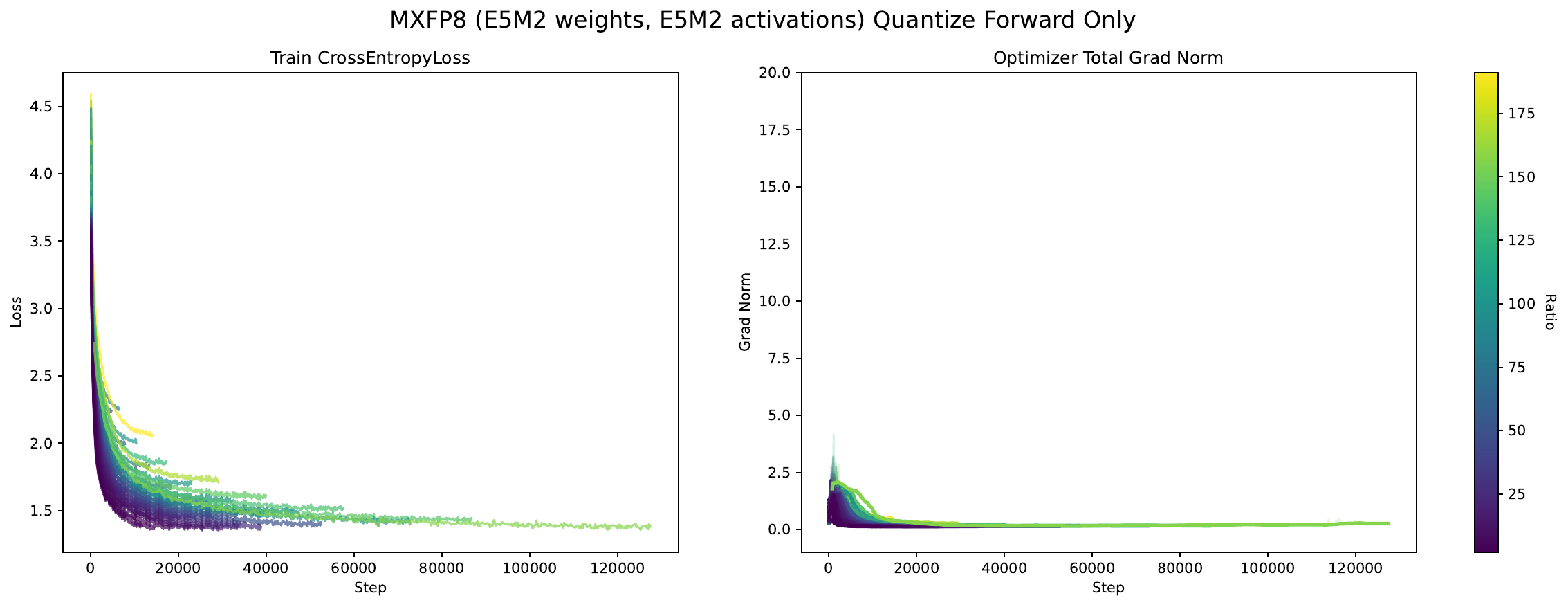} 
    \caption{Train loss and gradient norm for weights: \texttt{MXFP8 E5M2}, Activation: \texttt{MXFP8 E5M2}.}
    \label{fig: }
  \end{subfigure}
  \caption{Train loss and gradient norm when quantizing only the forward pass.}
  \label{fig: stable-quantize-fwd-only}
\end{figure}

\begin{figure}[ht]
  \centering
  \begin{subfigure}[t]{\textwidth}
    \centering
    \includegraphics[width=\linewidth]{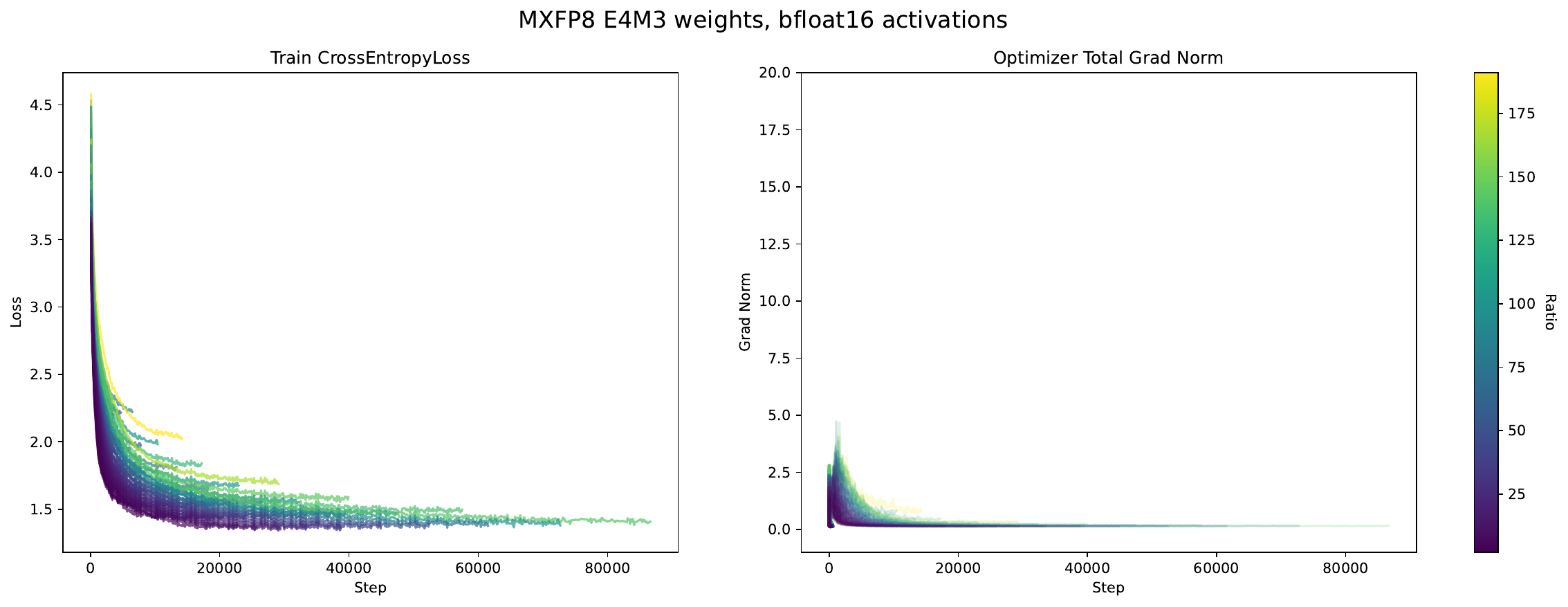} 
    \caption{Train loss and gradient norm for \texttt{MXFP8 E4M3}-\texttt{MXFP8 E4M3}.}
    \label{fig: scaling-law}
  \end{subfigure}
  \hfill
  \begin{subfigure}[t]{\textwidth}
    \centering
    \includegraphics[width=\linewidth]{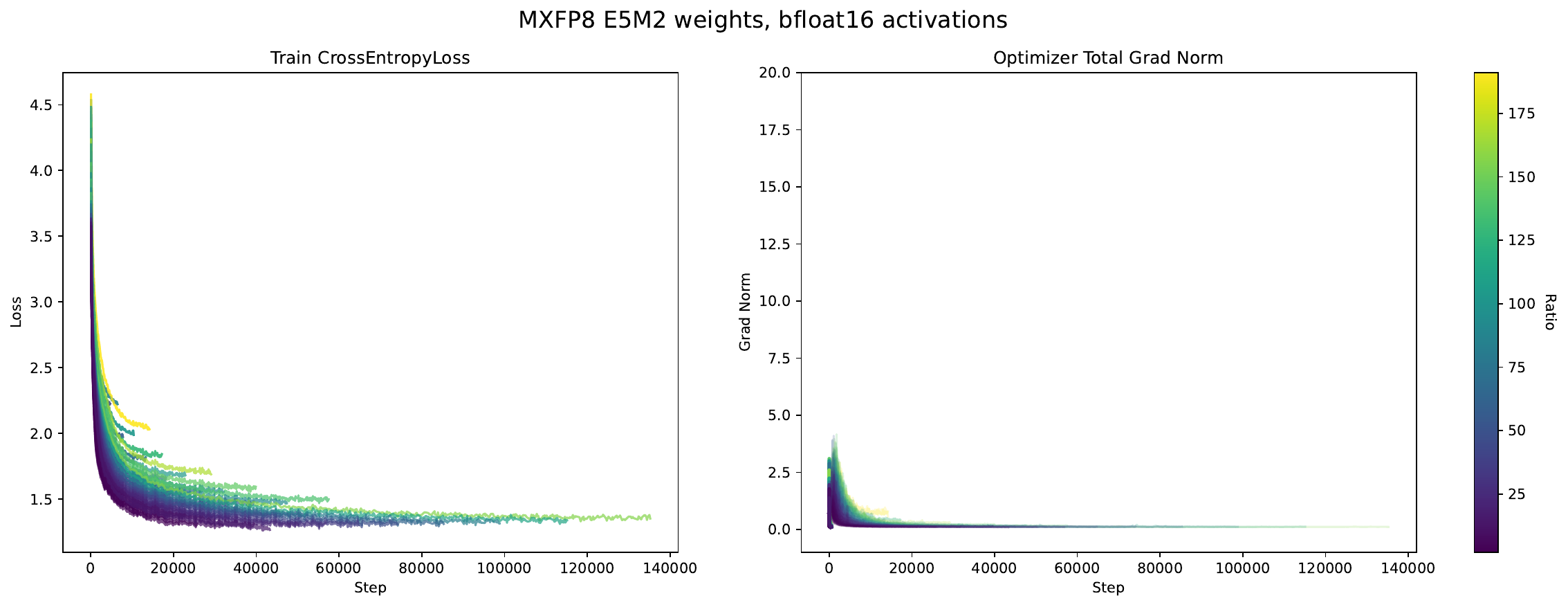} 
    \caption{Train loss and gradient norm for \texttt{MXFP8 E5M2}-\texttt{MXFP8 E5M2}.}
    \label{fig: scaling-law}
  \end{subfigure}
  \caption{Train loss and gradient norm when activations are kept in high precision (bfloat16).}
  \label{fig: stable-high-activations}
\end{figure}

\clearpage

\section{Details of Language Model Training} All models are trained on the Fineweb-Edu dataset~\citep{penedo2024the} using the Olmo codebase~\citep{groeneveld2024olmo}, with the longest runs trained on 35B tokens and the shortest runs corresponding to models trained on 301M tokens.  Models were trained with a learning rate schedule with a linear warmup starting at 2e-5 increasing to 2e-4, followed by cosine decay back to 2e-5~\citep{porian2025resolvingdiscrepanciescomputeoptimalscaling}.  Training runs that involved using MX precision formats were done performed using MX Pytorch Emulation Library~\citep{mx_library}.

\begin{table}[h]
\centering
\begin{tabular}{ll}
\toprule
\textbf{Parameter} & \textbf{Value} \\
\midrule
\textit{n} & 6–24 for small models \\
Number of heads & \textit{n} \\
Head dimension & 64 \\
MLP hidden multiplier & 4 \\
Depth & \textit{n} \\
Context length & 512 \\
Activation & GeLU \\
Positional encoding & RoPE \\
Biases & False \\
Normalization & PyTorch Layernorm \\
QK normalization & True \\
Tokenizer & Llama2 \\
\bottomrule \\
\end{tabular}
\caption{Model parameters used for training.}
\label{tab:model-params}
\end{table}

\clearpage

\section{Validation Losses in Language Models Quantizing}

\Cref{tab:MXFP8_val_loss_table} and continued in~\ref{tab:MXFP8_val_loss_table2} shows validation losses for all models with mitigations applied (quantization only in the forward pass, or activations in high precision), trained using our at different FLOP budgets relative to bfloat16 baseline.

\begin{table}[h]
\centering
\begin{tabular}{l c c c c c c}
\toprule
\multirow{2}{*}{\diagbox{\textbf{D/N}}{\textbf{formats}}} &  & \textbf{bfloat16} & \textbf{E4M3} & \textbf{E5M2} & \textbf{E4M3} & \textbf{E5M2} \\
& & \textbf{bfloat16} & \textbf{bfloat16}   & \textbf{bfloat16}   & \textbf{ E4M3} & \textbf{E5M2} \\
\midrule
\textbf{87.35} & \multirow{8}{*}{2e+17} & \cellcolor{gray!10}1.1522 & -0.027 & -0.027 & -0.027 & -0.012 \\
\textbf{46.99}  &  & \cellcolor{gray!10}1.1084 & 0.002 & 0.007 & 0.007 & 0.012 \\
\textbf{26.897}  &  & \cellcolor{gray!10}1.1011 & 0.004 & 0.001 & 0.004 & 0.009 \\
\textbf{16.06}  &  & \cellcolor{gray!10}1.0956 & -0.001 & 0.014 & 0.004 & 0.009 \\
\textbf{9.92}  &  & \cellcolor{gray!10}1.0971 & 0.003 & 0.003 & 0.008 & 0.013 \\
\textbf{6.30}  &  & \cellcolor{gray!10}1.0950 & 0.0 & -0.005 & -0.01 & 0.01 \\
\textbf{4.10}  &  & \cellcolor{gray!10}1.1042 & 0.001 & -0.006 & 0.006 & 0.006 \\
\textbf{2.73}  &  & \cellcolor{gray!10}1.1255 & -0.001 & -0.004 & 0.004 & 0.019 \\
\midrule
\textbf{191.02}  & \multirow{10}{*}{4.37e+17} & \cellcolor{gray!10}1.030 & 0.005 & 0.0 & 0.010 & 0.01 \\
\textbf{102.78}  &  & \cellcolor{gray!10}1.0464 & -0.016 & 0.036 & -0.011 & -0.021 \\
\textbf{58.81}  &  & \cellcolor{gray!10}0.9898 & 0.005 & 0.005 & 0.005 & 0.015 \\
\textbf{35.14}  &  & \cellcolor{gray!10}0.9806 & -0.001 & 0.004 & 0.004 & 0.009 \\
\textbf{21.70}  &  & \cellcolor{gray!10}0.9765 & 0.003 & 0.003 & 0.003 & 0.013 \\
\textbf{13.78}  &  & \cellcolor{gray!10}0.9717 & 0.003 & 0.003 & 0.003 & 0.008 \\
\textbf{8.97}  &  & \cellcolor{gray!10}0.9732 & 0.002 & 0.002 & 0.002 & 0.012 \\
\textbf{5.97}  &  & \cellcolor{gray!10}2.3174 & 0.303 & 0.843 & 2.763 & 1.237 \\
\textbf{4.05}  &  & \cellcolor{gray!10}0.9839 & 0.001 & 0.006 & 0.006 & 0.006 \\
\textbf{2.80}  &  & \cellcolor{gray!10}0.9949 & 0.0  & 0.0 & 0.0 & 0.005 \\
\midrule
\textbf{128.62}  & \multirow{10}{*}{9.56e+17} & \cellcolor{gray!10}0.9198 & 0.0 & 0.0 & 0.005 & 0.015 \\
\textbf{76.84}  &  & \cellcolor{gray!10}0.9052 & 0.0 & 0.005 & 0.005 & 0.015 \\
\textbf{47.46}  &  & \cellcolor{gray!10}0.8969 & 0.002 & 0.003 & 0.003 & 0.008 \\
\textbf{30.14}  &  & \cellcolor{gray!10}0.8894 & 0.001 & 0.001 & 0.006 & 0.011 \\
\textbf{19.62}  &  & \cellcolor{gray!10}0.8846 & 0.0 & 0.005 & 0.005 & 0.01 \\
\textbf{13.05}  &  & \cellcolor{gray!10}0.8879 & 0.002 & 0.002 & 0.002 & 0.012 \\
\textbf{8.86}  &  & \cellcolor{gray!10}0.8849 & 0.0 & 0.005 & 0.005 & 0.005 \\
\textbf{6.13}  &  & \cellcolor{gray!10}0.8882 & 0.002 & 0.002 & 0.002 & 0.007 \\
\textbf{4.31}  &  & \cellcolor{gray!10}0.8933 & 0.002 & 0.002 & 0.002 & 0.007 \\
\textbf{3.08}  &  & \cellcolor{gray!10}0.8961 & 0.004 & 0.004 & 0.004 & 0.009 \\
\textbf{2.24}  &  & \cellcolor{gray!10}0.9059 & -0.001 & 0.004 & 0.064 & 0.004 \\
\midrule
\textbf{168.03}  &  \multirow{10}{*}{2.09e+18} & \cellcolor{gray!10}0.8546 & 0.0 & 0.005 & 0.005 & 0.015 \\
\textbf{103.78}  &  & \cellcolor{gray!10}0.8430 & 0.002 & 0.002 & 0.187 & 0.012 \\
\textbf{65.91}  &  & \cellcolor{gray!10}0.8335 & 0.001 & 0.001 & 0.001 & 0.011 \\
\textbf{42.896}  &  & \cellcolor{gray!10}0.8258 & -0.001 & 0.004 & 0.004 & 0.009 \\
\textbf{28.54}  &  & \cellcolor{gray!10}0.8242 & 0.001 & 0.001 & 0.001 & 0.011 \\
\textbf{19.37}  &  & \cellcolor{gray!10}0.8200 & 0.0 & 0.0 & 0.0 & 0.005 \\
\textbf{13.399}  &  & \cellcolor{gray!10}0.8197 & 0.0 & 0.0 & 0.0 & 0.005 \\
\textbf{9.428}  &  & \cellcolor{gray!10}0.8187 & 0.001 & 0.001 & 0.001 & 0.006 \\
\textbf{6.74}  &  & \cellcolor{gray!10}0.8192 & 0.001 & 0.001 & 0.001 & 0.006 \\
\textbf{4.89}  &  & \cellcolor{gray!10}0.8215 & 0.003 & 0.006 & 0.003 & 0.003 \\
\textbf{2.02}  &  & \cellcolor{gray!10}0.8327 & 0.002 & 0.002 & 0.002 & 0.002 \\
\bottomrule \\
\end{tabular}
\caption{  Validation loss table, with separate columns for various weight and activation precisions. For the last 2 columns, we quantize only the forward pass. The second column indicates the total FLOP count used for those values of tokens-to-parameter ratios ($D/N$). Values are shown as differences with respect to bfloat16 baseline (lower is better).}
\label{tab:MXFP8_val_loss_table}
\end{table}

\begin{table}[h]
\centering
\begin{tabular}{l c c c c c c}
\toprule
\multirow{2}{*}{\diagbox{\textbf{D/N}}{\textbf{formats}}} &  & \textbf{bfloat16} & \textbf{E4M3} & \textbf{E5M2} & \textbf{E4M3} & \textbf{E5M2} \\
& & \textbf{bfloat16} & \textbf{bfloat16}   & \textbf{bfloat16}   & \textbf{ E4M3} & \textbf{E5M2} \\
\midrule
\textbf{144.14} & \multirow{10}{*}{4.57e+18} & \cellcolor{gray!10}0.794 & 0.001 &  0.006 & 0.006 & 0.011 \\
\textbf{93.81} & & \cellcolor{gray!10}0.784 & 0.001 & 0.001 & 0.001 & 0.011 \\
\textbf{62.41} & & \cellcolor{gray!10}0.780 & 0.0 & 0.005 & 0.005 & 0.01 \\
\textbf{42.37} & & \cellcolor{gray!10}0.774 & 0.001 & 0.001 & 0.001 & 0.006 \\
\textbf{29.30} & & \cellcolor{gray!10}0.772 & -0.002 & 0.003 & 0.003 & 0.003 \\
\textbf{14.74} & & \cellcolor{gray!10}0.767 & -0.002 & 0.003 & 0.003 & 0.003 \\
\textbf{10.70} & & \cellcolor{gray!10}0.766 & -0.001 & 0.004 &  -0.001 & 0.004 \\
\textbf{7.87} & & \cellcolor{gray!10}0.766 & -0.001 & 0.004 &  -0.001 & 0.004 \\
\textbf{4.42} & & \cellcolor{gray!10}0.769 & 0.001 & 0.001 & 0.001 & 0.006 \\
\textbf{3.37} & & \cellcolor{gray!10}0.772 &  -0.002 & 0.003 &  0.003 & 0.003 \\
\textbf{2.60} & & \cellcolor{gray!10}0.775 &  0.0 & 0.0 & 0.0 & 0.005 \\
\textbf{2.02} & & \cellcolor{gray!10}0.779 &  0.001 & 0.001 & 0.001 & 0.001 \\
\midrule
\textbf{136.47458} & \multirow{10}{*}{1e+19} & \cellcolor{gray!10}0.748 & 0.002 & 0.002 & 0.002 & 0.002 \\
\textbf{92.646} & & \cellcolor{gray!10} 0.741 & -0.001 & 0.004 & 0.004 & 0.009 \\
\textbf{64.075} & & \cellcolor{gray!10} 0.736 & -0.001 & 0.004 & 0.004 & 0.009 \\
\textbf{45.084} & & \cellcolor{gray!10} 0.731 & -0.001 & 0.004 & 0.004 & 0.009 \\
\textbf{32.233} & & \cellcolor{gray!10} 0.728 & 0.002 & 0.002 & 0.002 & 0.007 \\
\textbf{23.391} & & \cellcolor{gray!10} 0.725 & 0.0 & 0.005 & 0.0 & 0.005 \\
\textbf{17.210} & & \cellcolor{gray!10} 0.724 & 0.001 & 0.001 & 0.001 & 0.006 \\
\textbf{12.826} & & \cellcolor{gray!10} 0.724 & 0.001 & 0.001 & 0.001 & 0.311 \\
\textbf{9.674} & & \cellcolor{gray!10} 0.723 & 0.002 & 0.002 & 0.002 & 0.002 \\
\textbf{7.38} & & \cellcolor{gray!10} 0.723 & 0.002 & 0.002 & 0.002 & 0.077 \\
\textbf{4.43} & & \cellcolor{gray!10} 0.727 & -0.002 & 0.003 & 0.003 & 0.003 \\
\textbf{2.75} & & \cellcolor{gray!10} 0.732 & -0.002 & 0.023 & 0.003 & 0.003 \\
\bottomrule \\
\end{tabular} 
\caption{ MXFP8 of the validation loss table, with separate rows for Weight and Activation precisions. For the last 2 columns, we quantize only the forward pass. The second column indicates the FLOP count used.} \label{tab:MXFP8_val_loss_table2}
\end{table}

\clearpage

\section{Additional Unstable Language Model Sweeps\label{app:additional_lang_sweeps}}  In~\Cref{fig: scaling-laws-unstable1} and~\Cref{fig: scaling-laws-unstable2} we show some other examples of weight/activation MX precision combinations we found to be unstable.  In general, we were not able to find any stable combinations of weights and activations in lower precision across the formats we tested.

\begin{figure}[ht]
  \centering
  \begin{subfigure}[t]{\textwidth}
    \centering
    \includegraphics[width=\linewidth]{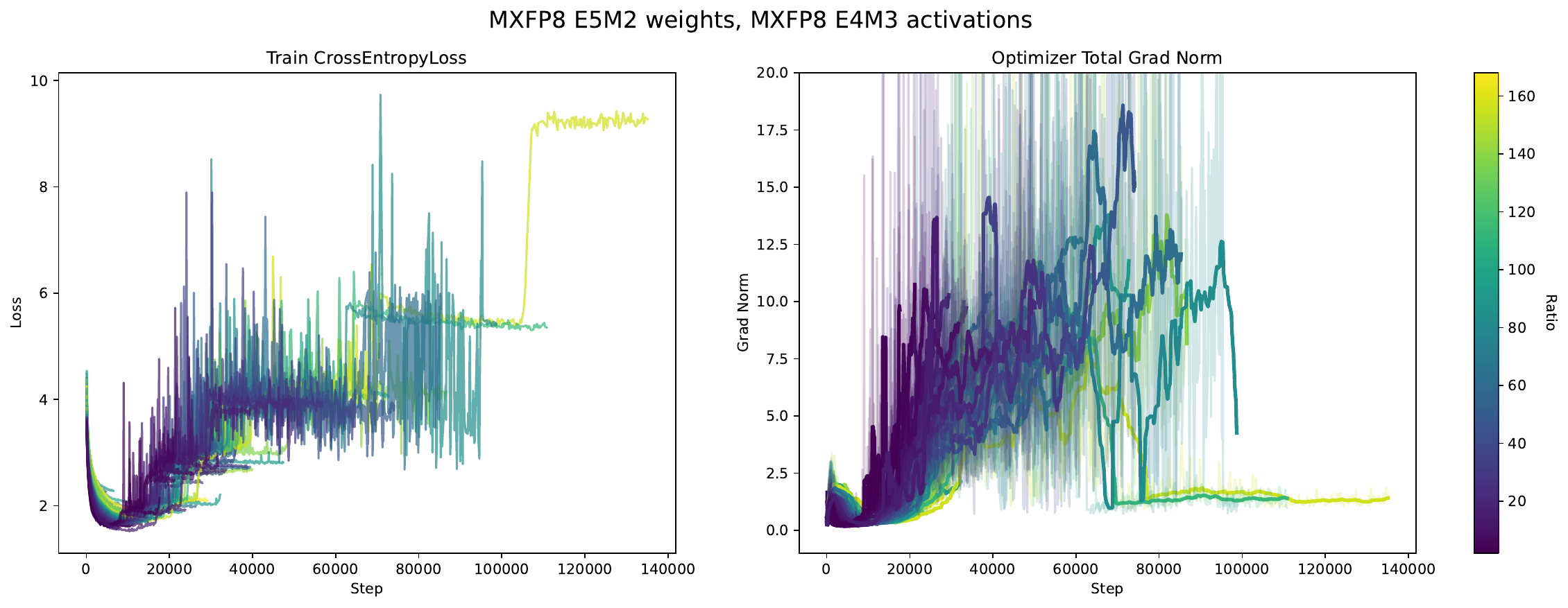} 
    \caption{Train loss and grad norm for \texttt{MXFP8 E5M2}-\texttt{MXFP8 E4M3}.}
    \label{fig: }
  \end{subfigure}
  \\
  \begin{subfigure}[t]{\textwidth}
    \centering
    \includegraphics[width=\linewidth]{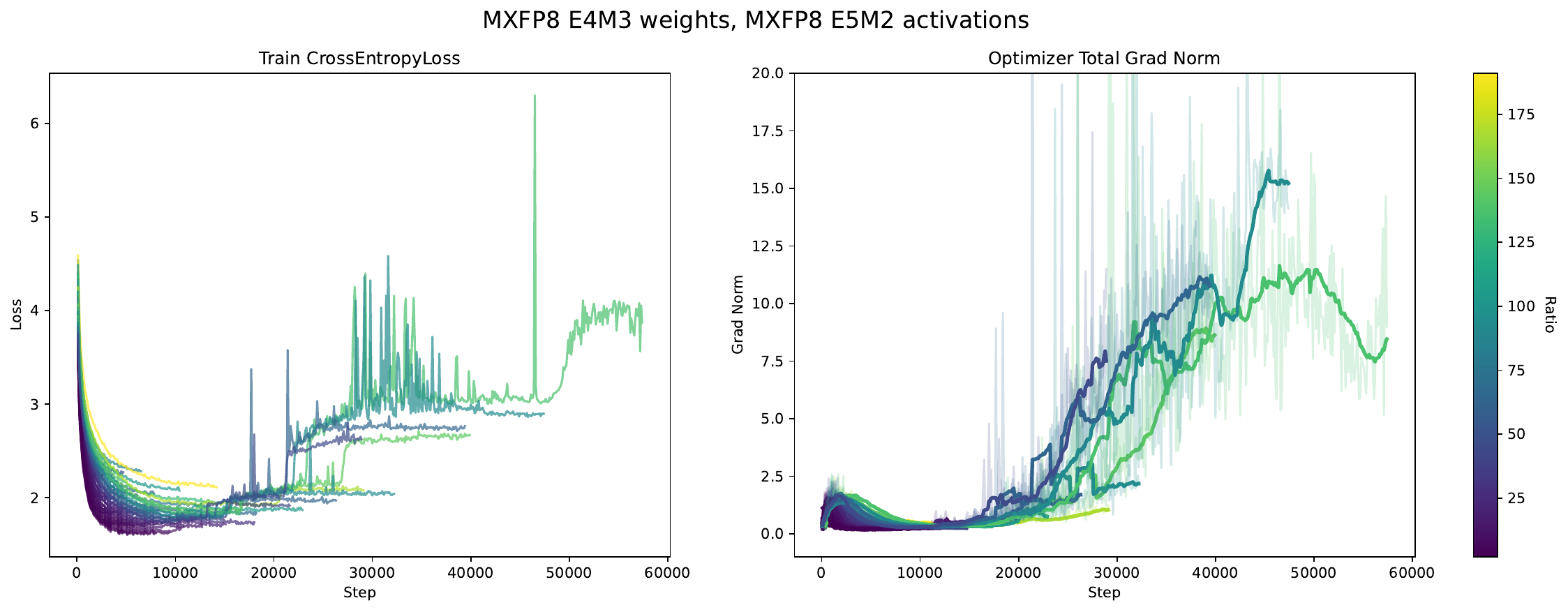} 
    \caption{Train loss and grad norm for \texttt{MXFP8 E4M3}-\texttt{MXFP E5M2}.}
    \label{fig: }
  \end{subfigure}
  \hfill
  \caption{ \textbf{Unstable} MXFP8 combinations of precision formats of weights and activations.}
  \label{fig: scaling-laws-unstable1}
\end{figure}

\begin{figure}[ht]
  \centering
  \begin{subfigure}[t]{\textwidth}
    \centering
    \includegraphics[width=\linewidth]{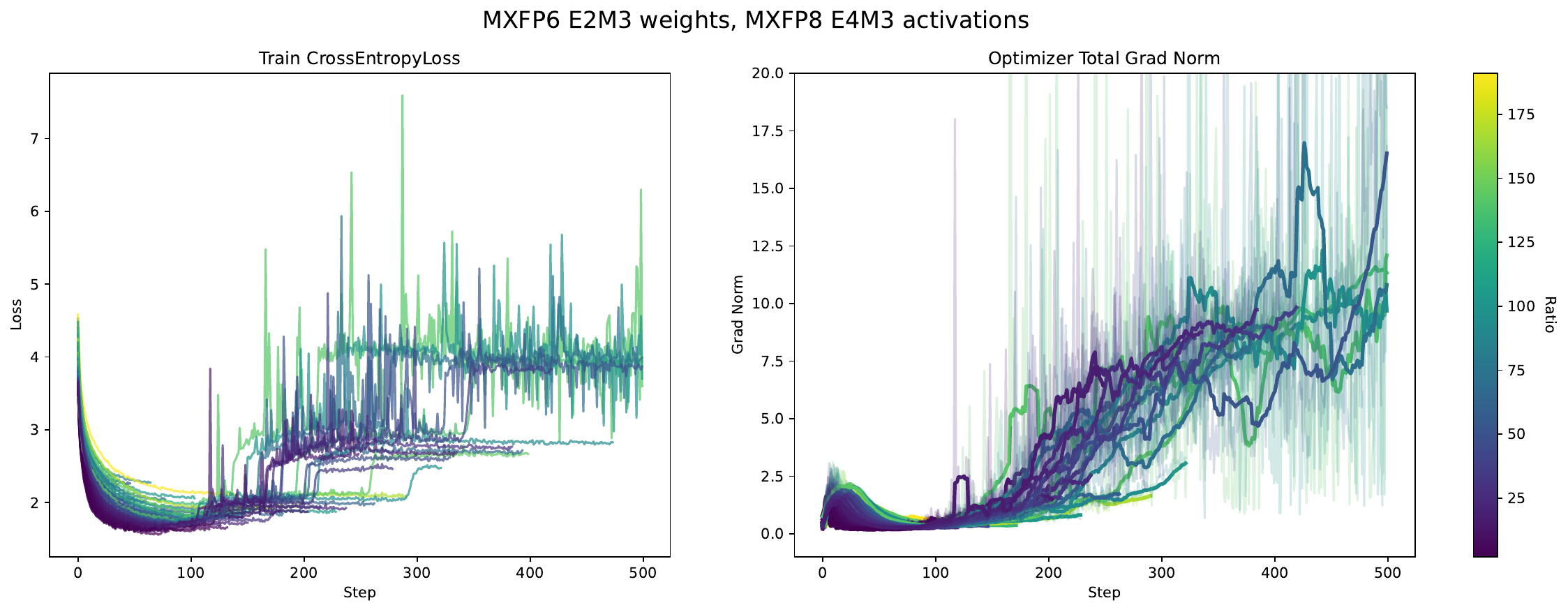} 
    \caption{ for \texttt{MXFP6 E2M3}-\texttt{MXFP8 E4M3}.}
    \label{fig: }
  \end{subfigure}
  \\
  \begin{subfigure}[t]{\textwidth}
    \centering
    \includegraphics[width=\linewidth]{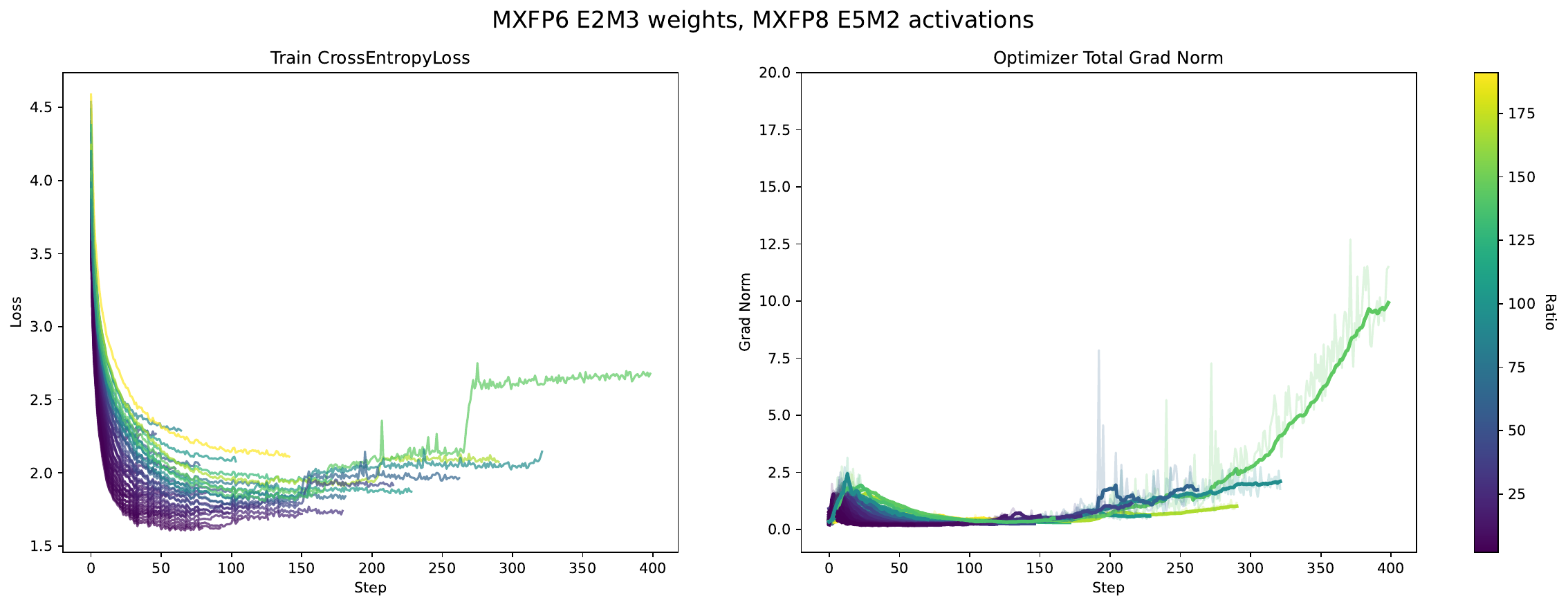} 
    \caption{Train loss and  for \texttt{MXFP6 E2M3}-\texttt{MXFP8 E5M2}.}
    \label{fig: }
  \end{subfigure}
  \hfill
  \caption{\textbf{Unstable} combinations of precision formats of weights and activations for MXFP6 weights.}
  \label{fig: scaling-laws-unstable2}
\end{figure}

\clearpage

\bibliographystyle{apalike}
\bibliography{bibliography}

\end{document}